\documentclass[lettersize,journal]{IEEEtran}
\usepackage{amsmath,amsfonts}
\usepackage{algorithm}
\usepackage{array}
\usepackage[caption=false,font=normalsize,labelfont=sf,textfont=sf]{subfig}
\usepackage{textcomp}
\usepackage{stfloats}
\usepackage{url}
\usepackage{verbatim}
\usepackage{graphicx}
\usepackage{cite}
\usepackage{soul}

\usepackage{hyperref}
\usepackage{url}
\usepackage{hyperref}
\usepackage{url}
\usepackage{booktabs}
\usepackage{array}
\usepackage{caption}
\usepackage{pifont}
\usepackage[table]{xcolor}
\usepackage{wrapfig}
\usepackage{multirow}
\usepackage{multicol}
\usepackage{amsmath, amssymb, amsthm, mathrsfs}
\usepackage{bm}
\usepackage{algorithm}
\usepackage[noend]{algpseudocode}
\usepackage{float}
\usepackage{enumitem}
\usepackage{tabularx}
\usepackage[utf8]{inputenc}
\usepackage{hyperref}
\definecolor{cite_color}{RGB}{119, 185, 0}  
\hypersetup{
    colorlinks=true,
    linkcolor=red,
    citecolor=cite_color,
    filecolor=mycustompurple,
    urlcolor=magenta
}

\definecolor{gray}{rgb}{0.5,0.5,0.5}
\definecolor{Gray}{gray}{0.93}

\def\CircleArrowright{\ensuremath{
  \rotatebox[origin=c]{310}{$\circlearrowright$}}}

\newcommand{\vlnbert}{VLN$\protect\CircleArrowright$BERT}
\newcommand{\halfcheck}{\ding{51}$\mkern-12mu\raisebox{0.4ex}{$\bm\smallsetminus$}$}
\newcommand{\etal}{\textit{et al.}}

\begin{document}

\title{Constraint-Aware Zero-Shot Vision-Language Navigation in Continuous Environments}

\author{Kehan Chen\thanks{$*$ These two authors contributed equally to this work.}$^{*}$,
Dong An$^*$,
Yan Huang,
Rongtao Xu,
Yifei Su,
Yonggen Ling,
Ian Reid,
Liang Wang\thanks{$^\dag$ Liang Wang is the corresponding author.}$^\dag$
\IEEEcompsocitemizethanks{
\IEEEcompsocthanksitem 
Kehan Chen, Yan Huang, Yifei Su, and Liang Wang are with New Laboratory of Pattern Recognition (NLPR), Institute of Automation, Chinese Academy of Sciences and School of Artificial Intelligence, University of Chinese Academy of Sciences, China. Email: 
kehan.chen@cripac.ia.ac.cn, yhuang@nlpr.ia.ac.cn, yifei.su@cripac.ia.ac.cn, wangliang@nlpr.ia.ac.cn.
\IEEEcompsocthanksitem 
Dong An, Rongtao Xu, and Ian Reid are with Mohamed bin Zayed University of Artificial Intelligence (MBZUAI), Abu Dhabi, UAE. Email: 
dong.an@mbzuai.ac.ae, rongtao.xu@mbzuai.ac.ae, ian.reid@mbzuai.ac.ae.
\IEEEcompsocthanksitem 
Yonggen Ling is with Robotics X, Tencent. Email: rolandling@tencent.com.
}}

\markboth{Journal of \LaTeX\ Class Files,~Vol.~14, No.~8, August~2021}%
{Shell \MakeLowercase{\textit{et al.}}:Constraint-Aware zero-shot Vision-Language Navigation in Continuous Environments}


\maketitle

\begin{abstract}
We address the task of  Vision-Language Navigation in Continuous Environments (VLN-CE) under the zero-shot setting. 
Zero-shot VLN-CE is particularly challenging due to the absence of expert demonstrations for training and minimal environment structural prior to guide navigation.
To confront these challenges, we propose a Constraint-Aware Navigator (CA-Nav), which reframes zero-shot VLN-CE as a sequential, constraint-aware sub-instruction completion process.
CA-Nav continuously translates sub-instructions into navigation plans using two core modules: the Constraint-Aware Sub-instruction Manager (CSM) and the Constraint-Aware Value Mapper (CVM).
CSM defines the completion criteria for decomposed sub-instructions as constraints and tracks navigation progress by switching sub-instructions in a constraint-aware manner.
CVM, guided by CSM's constraints, generates a value map on the fly and refines it using superpixel clustering to improve navigation stability.
CA-Nav achieves the state-of-the-art performance on two VLN-CE benchmarks, surpassing 
the previous best method by 12\% and 13\% in Success Rate on the validation unseen splits of R2R-CE and RxR-CE, respectively. 
Moreover, CA-Nav demonstrates its effectiveness in real-world robot deployments across various indoor scenes and instructions\footnote{Project page:~\href{https://chenkehan21.github.io/CA-Nav-project/}{https://chenkehan21.github.io/CA-Nav-project/}}.
\end{abstract}


\section{Introduction}
\IEEEPARstart{V}{ISION}-Language Navigation (VLN) is a fundamental task in Embodied AI. 
It requires the agent to navigate in novel environments according to natural language instructions~\cite{anderson2018vision}.
Early efforts primarily focused on discrete environments, leveraging structural priors such as connectivity graphs to guide navigation~\cite{chen2022duet, an2024etpnav}. 
However, this setting diverges from real-world scenarios, where structural prior like connectivity graph is typically unavailable, and collecting expert demonstrations for training is costly.
The lack of structural prior and expert prior thus poses critical challenges for advancing practical VLN applications.
Recent studies have addressed these limitations along two orthogonal dimensions, as illustrated in Figure~\ref{fig:zs-vlnce-challenges}.
First, Vision-Language Navigation in Continuous Environments (VLN-CE)~\cite{krantz_vlnce_2020} has gained traction by allowing agents to navigate freely within 3D environments via navigation meshes, which alleviates the reliance on structural prior like connectivity graphs.
Second, the emergence of Vision-Language Models (VLMs)~\cite{li2023blip2} and Large Language Models (LLMs)~\cite{openai2024gpt4} has enabled zero-shot VLN, which aims to perform navigation tasks using foundation models without relying on expert-annotated training data. 
\begin{center}
\begin{figure}[t]
\centering
\includegraphics[width=0.45\textwidth]{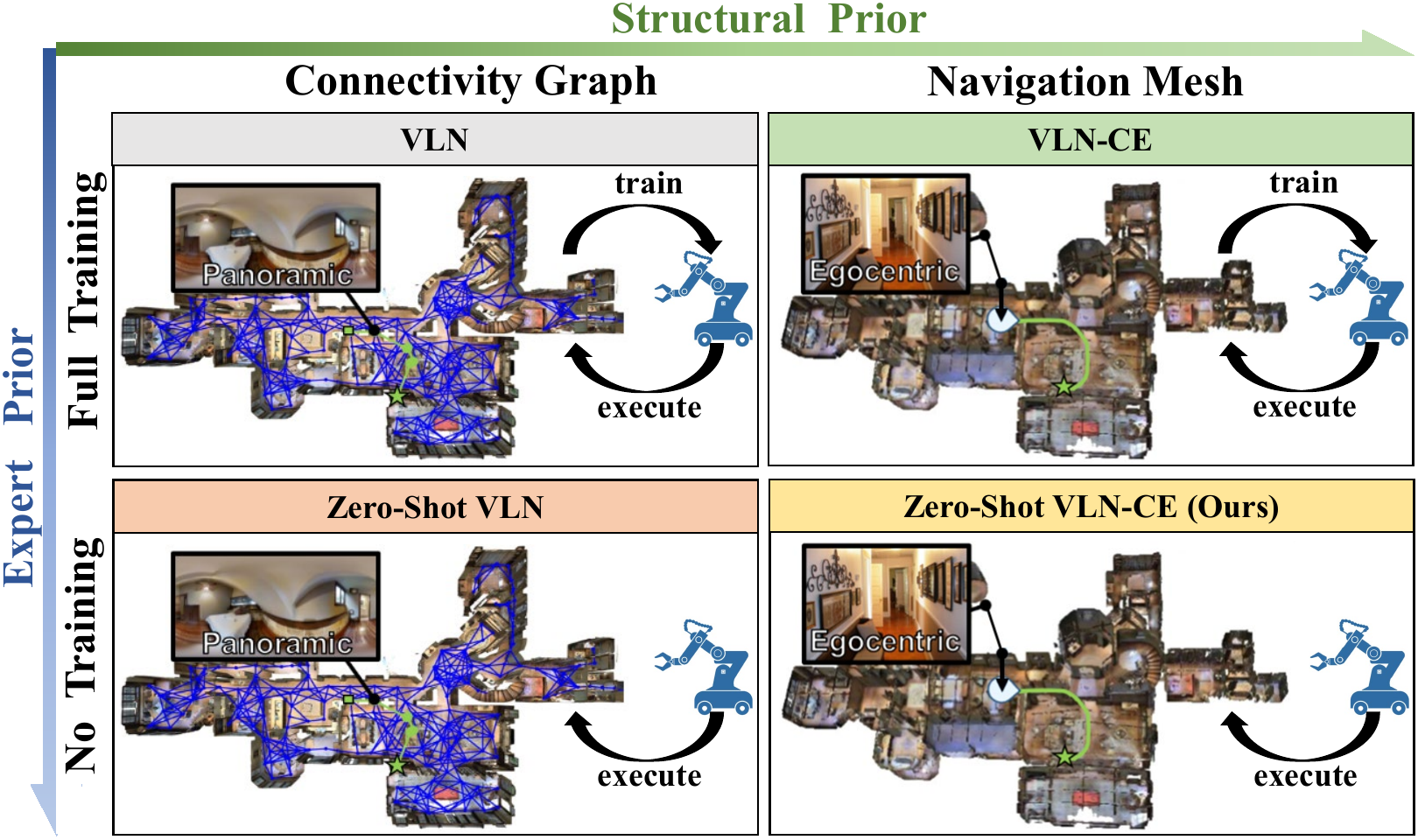}
\caption{
The zero-shot VLN-CE task (bottom right) addresses the dual challenge of prior scarcity along two orthogonal dimensions: the absence of expert demonstrations for training and limited structural prior to guide navigation.
}
\label{fig:zs-vlnce-challenges}
\end{figure}
\vspace{-2em}
\end{center}

Despite promising progress in addressing the scarcity of expert prior and structural prior individually, the combination of these two research lines - zero-shot VLN-CE - remains largely unexplored.
A potential approach is to adapt existing zero-shot VLN methods to continuous environments. 
Existing zero-shot VLN methods~\cite{zhou2024navgpt,chen2024mapgpt, qiao2024opennav} follow a text-based prompt paradigm, converting visual observations and navigation history into text, which is then combined with the full instruction and input into an LLM to infer the next action.
However, such an adaptation faces two major challenges and shows substantial performance degradation.
First, continuous environments expand the state space, making it difficult for the agent to accurately track navigation progress and determine which part of the instruction is being executed.
Existing methods that input the full instruction into the LLM might overlook the importance of monitoring the completion status of sub-instructions.
Second, converting visual observations into text often leads to losing visual details and environmental structures~\cite{wunderlich2021landmark}, impairing the agent's spatial understanding and path planning.

In light of the above, we propose Constraint-aware Navigator (CA-Nav), a new approach to tackle the challenges of zero-shot VLN-CE. 
CA-Nav reframes VLN-CE as a sequential, constraint-aware sub-instruction completion process.
Within each episode, the Constraint-Aware Sub-instruction Manager (CSM) decomposes the given instruction into sub-instructions and dynamically switches between them by assessing whether the relevant constraints are met. 
Meanwhile, the Constraint-aware Value Mapper (CVM) constructs and continuously updates a value map based on current constraints and observations, capturing both visual details and spatial layouts. 
Using the value map generated by CVM, CA-Nav formulates navigation plans, which are executed by classical control algorithms.
This iterative process guides the agent to complete each sub-instruction until the episode terminates.

\begin{figure*}[t]
  \centering
  \includegraphics[width=0.9\textwidth]{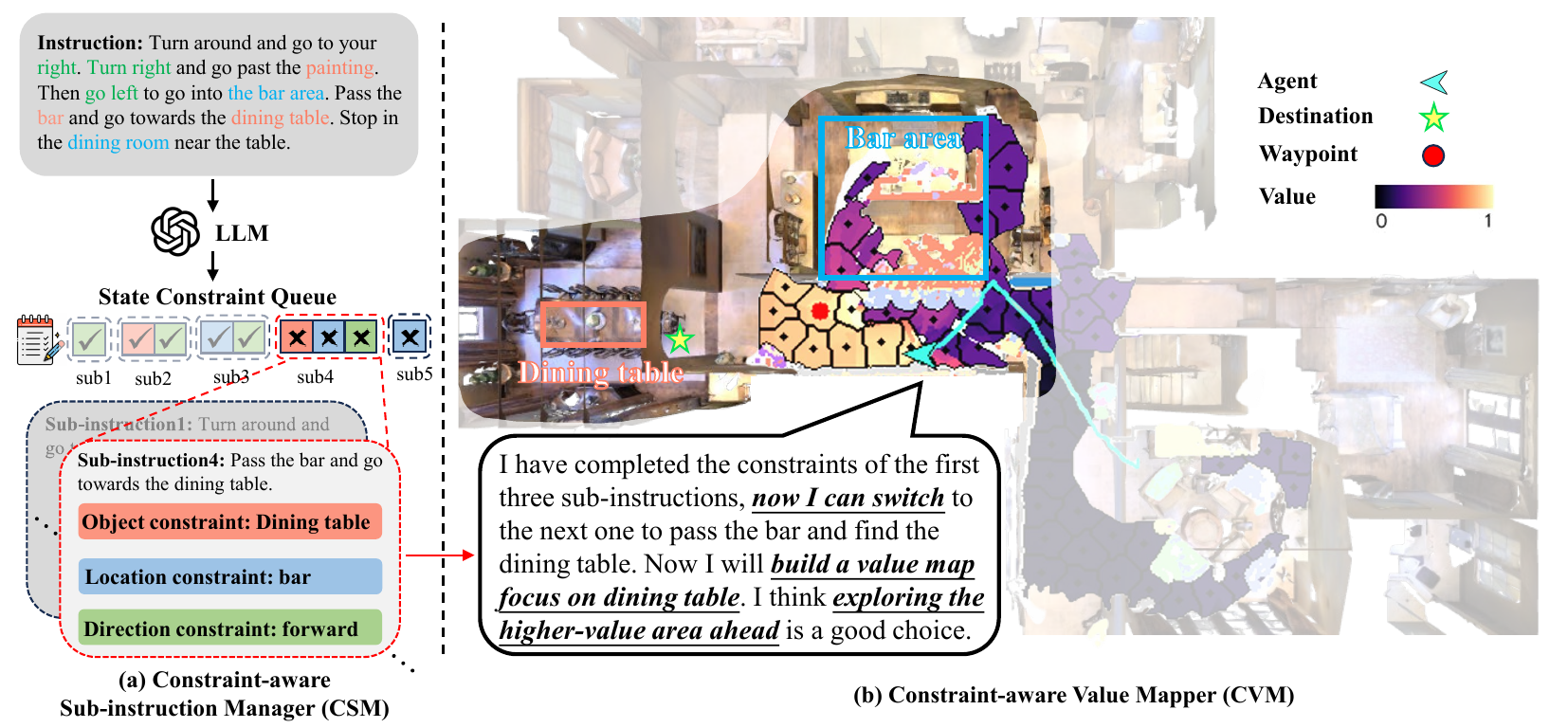}
  \caption{
  Illustration of the proposed CA-Nav.
  (a) The Constraint-aware Sub-instruction Manager decomposes the instruction into a sequence of sub-instructions and identifies \textcolor[rgb]{0.98, 0.58, 0.5}{object constraints}, \textcolor[rgb]{0.61, 0.76, 0.90}{location constraints} and \textcolor[rgb]{0.66, 0.82, 0.55}{direction constraints} for each of them. 
  (b) During navigation, a Constraint-aware Value Mapper builds a value map based on the landmark prompt provided by CSM and uses the superpixel clustering method to segment it into regions. It switches sub-instructions in a constraint-aware manner and chooses the most promising region's geometric center as waypoints.}
  \label{fig:teaser}
\end{figure*}

As shown in Figure \ref{fig:teaser} (a), at the start of an episode, CSM prompts the LLM to decompose the instruction into sub-instructions and generate constraints for completing each one.
During navigation, CSM actively monitors the fulfillment of these constraints and switches to the next sub-instruction once the current constraints are met.
To support this process, we leverage VLMs~\cite{liu2023grounding, li2023blip2} to detect various constraints, including landmark detection, location recognition, and direction estimation.
These constraints are composed to cover diverse sub-instruction expressions, ensuring flexibility and robustness in handling different instructions.
\captionsetup[figure]{aboveskip=1pt, belowskip=0pt}

Building on the sub-instructions identified by CSM, the next challenge is grounding them in the map alongside visual details for improved spatial understanding and navigation planning. 
We address this with CVM, which evaluates the potential of each step's observations in satisfying the current constraint. 
As depicted in Figure~\ref{fig:teaser} (b), CVM builds a map to capture both the semantics and spatial layout of the environment. 
Using a VLM, it calculates the similarity between current observations and landmarks associated with task constraints, generating a constraint-aware value map projected onto the ground plane. 
To enhance the accuracy and stability of navigation, superpixel clustering~\cite{achanta2012slic} is applied to refine the map, reducing noise and maintaining coherence within regions. 
This enables the agent to select waypoints from high-value regions, ensuring that navigation aligns with both task constraints and environmental understanding.
Finally, classical control algorithms~\cite{sethian1999fast} are employed for waypoint-reaching navigation.

Experiments in both simulation and real-world demonstrate the effectiveness of CA-Nav. 
On two VLN-CE benchmarks under the zero-shot setting, CA-Nav achieves state-of-the-art performance, surpassing the previous best methods by 12\% on R2R-CE and 13\% on RxR-CE in terms of success rate on the validation unseen split.
Notably, CA-Nav delivers approximately 3 times faster response times and a 95\% cost reduction compared to counterparts. 
Furthermore, real-world robot deployments validate CA-Nav's potential for practical applications, showcasing its effectiveness across open-vocabulary instructions and various indoor environments.

The rest of the paper is organized as follows:
In \S~\ref{related work}, we provide a brief review of the related work.
\S~\ref{method} outlines the task setup for zero-shot Vision-Language Navigation in Continuous Environments and details the proposed method.
Experimental results and ablation studies are presented in \S~\ref{experiments}.
Finally, we conclude this work in \S~\ref{conclusion}.

\vspace{-6pt}
\section{Related Work}
\label{related work}
\subsection{Vision-Language Navigation.} Vision-Language Navigation (VLN) has garnered significant attention in recent years.
Early VLN approaches leveraged LSTM models~\cite{LSTM} for cross-modal alignment and navigation policy learning~\cite{anderson2018vision}.
Building upon these methods, various techniques have been proposed, including progress monitoring~\cite{ma2019self, ma2019regretful}, backtracking mechanisms~\cite{ke2019tactical}, and reinforcement learning~\cite{wang2019reinforced, wang2018look}.
Some approaches focus on extracting specific information from both instructions and observations, such as rooms~\cite{gao2021room, qi2021road}, objects~\cite{qi2020object, moudgil2021soat, qi2021road}, scenes~\cite{lin2021scene, moudgil2021soat}, and actions~\cite{qi2020object}.
In addition, supplementary information, such as navigation history~\cite{chen2021history}, neighboring views~\cite{an2021neighbor}, and topological graphs~\cite{chen2022duet, an2024etpnav}, has also been explored to improve navigation performance.

More recently, transformer-based methods have become the predominant approach in VLN due to their superior capability in cross-modal alignment~\cite{vaswani2017attention, qi2021road, moudgil2021soat, hong2021vln}.
This shift has also led to the adoption of pre-training strategies~\cite{hao2020prevalent, majumdar2020improving}, which incorporate auxiliary tasks~\cite{Qiao2022HOP, qiao2023hop+, hong2024only} to learn general representations that can be applied to the VLN task.
Furthermore, a significant body of research has focused on augmenting VLN datasets to enhance the agent's generalization ability in novel environments.
Some studies have leveraged internet data such as YouTube videos~\cite{guhur2021airbert, lin2023learning}, while others have synthesized new scenes~\cite{li2022envedit, chen2022learning, wang2023scalevln, he2024frequency, li2024panogen} and instructions~\cite{fried2018speaker, he2021landmark} to diversify training data and improve model robustness.

Although significant progress has been made, these methods overlook low-level control in continuous environments, particularly in scenarios where predefined navigation graphs are unavailable.
As a result, their performance drops significantly when VLN methods are applied to real-world settings~\cite{anderson2020sim}. 

\vspace{-6pt}
\subsection{VLN in Continuous Environments.}
Since the assumption of a predefined navigation graph does not accurately reflect the challenges in the real-world, Krantz\etal~\cite{krantz_vlnce_2020} introduce Vision-Language Navigation in Continuous Environments (VLN-CE).
Early VLN-CE approaches use GRU networks~\cite{2014Empirical} to align observations and instructions, and then directly learn low-level actions in an end-to-end manner~\cite{krantz_vlnce_2020, krantz2021waypoint}.
In order to improve the agent's instruction following capability, Raychaudhuri\etal~\cite{raychaudhuri2021law} uses the action of moving towards the ground-truth waypoint as supervision to improve language-aligned waypoint prediction.
Additionally, several studies incorporate semantic map for spatial reasoning, as these maps capture the layout of the environment and enhance waypoint prediction~\cite{chen2022weakly, wang2023gridmm, he2023mlanet}.

Hong\etal~\cite{hong_2022_bridging_the_gap} introduces a waypoint predictor trained on the navigation graphs from Matterport 3D dataset~\cite{chang2017matterport3d}.
When inputting current panoramic observation, the waypoint predictor outputs candidate waypoints.
After generating candidate waypoints by leveraging the waypoint predictor, the VLN-CE task is then simplified as a cross-modal waypoint selection problem which is similar to the VLN task.
It bridges the gap between discrete environments and continuous environments to some extent and brings an emerging trend of panoramic waypoint-based approaches~\cite{krantz2022sim2sim, an2023bevbert, an2024etpnav, wang2024sim}.

However, these methods rely heavily on annotated trajectories, which demand substantial human effort to create. 
Therefore, we aim to develop a zero-shot approach for the VLN-CE task relying on foundation models.

\subsection{Foundation Models for Robotic Navigation.} Foundation models are those trained on broad data that can be adapted to a wide range of downstream tasks including LLMs~\cite{gpt3, openai2024gpt4, touvron2023llama} and VLMs~\cite{radford2021clip, li2023blip2, liu2023grounding}. 
Their strengths in reasoning, task planning, and multi-modal understanding make them promising for robotic navigation~\cite{zhang2024navid, zheng2024towards, zhou2024navgpt2, lin2024navcot, chen2024aoplanner}.

Recently, zero-shot VLN methods~\cite{long2023discuss, zhou2024navgpt, zhan2024mc} have emerged, they use VLMs to describe observations and LLMs to make step-by-step decisions in discrete environments. 
Specifically, DiscussNav~\cite{long2023discuss} employs multiple GPT-4 experts to discuss the current observations, status, and instructions before moving; NavGPT~\cite{zhou2024navgpt} utilizes GPT-4 to process descriptions of visual observations and navigation history before navigating; MapGPT~\cite{chen2024mapgpt} converts a topological map into prompts and then uses GPT-4 for navigation; MC-GPT~\cite{zhan2024mc} also introduces a topological map to store history and leverages chain-of-thought (CoT)~\cite{wei2023cot} for navigation reasoning.

However, few studies focus on zero-shot VLN-CE~\cite{chen2023A2Nav, long2024instnav, qiao2024opennav}.
A$^2$Nav~\cite{chen2023A2Nav} attempts to decompose instructions into action-specific object navigation sub-tasks.
However, it's not truly training-free as it relies on room region bounding box annotations from HM3D~\cite{hm3d} to collect data for training five action-specific navigators.
OpenNav~\cite{qiao2024opennav} aims to develop a navigation system that leverages open-source LLMs and VLMs.
Although it focuses on continuous environments, it relies on the pre-trained waypoint predictor~\cite{hong_2022_bridging_the_gap} to provide candidate waypoints.
The navigation decisions are based on CoT reasoning, which also analyzes the descriptions of visual observations, similar to approaches for discrete environments like NavGPT and MC-GPT.
InstructNav~\cite{long2024instnav} utilizes GPT-4V~\cite{yang2023gpt4v} and panoramic observations to build an intuition value map for heading evaluation and constructs additional value maps based on hand-designed rules.
At each decision step, it calls GPT-4~\cite{openai2024gpt4} to plan the next navigation sub-task and then uses the multi-sourced value maps to select the waypoint.

In contrast, we introduce a novel training-free method, CA-Nav, which enables sub-instruction switching in a constraint-aware manner based on egocentric views.
In contrast to InstructNav, our approach focuses on a setting where the agent relies solely on egocentric observations. 
Additionally, CA-Nav does not require calling LLMs at every decision step, making it faster and more cost-efficient than InstructNav.

\vspace{-6pt}
\section{Method}
\label{method}
\textbf{Problem Definition.}
\label{problem definition}
We address the zero-shot VLN-CE task~\cite{krantz_vlnce_2020}, where the agent navigates to a destination following natural language instructions. 
Unlike methods like A$^2$Nav~\cite{chen2023A2Nav}, which are trained on datasets for specific navigation skills, our approach only utilizes foundation models for decision-making.
To better align with real-world scenarios, we adopt an egocentric setup, which differs from works like InstructNav~\cite{long2024instnav}.
Concretely, the agent is equipped with odometry and an egocentric RGB-D camera with a 79$^\circ$ Horizontal Field of View (HFOV).
It can perform low-level actions such as MOVE FORWARD (0.25m), TURN LEFT/RIGHT (30$^\circ$), and STOP.
An episode is considered successful if the agent stops within a certain distance from the target. 

\vspace{-6pt}
\subsection{Method Overview}
\begin{figure*}[t]
  \centering
  \includegraphics[width=0.8\textwidth, trim=0 10 0 10, clip]{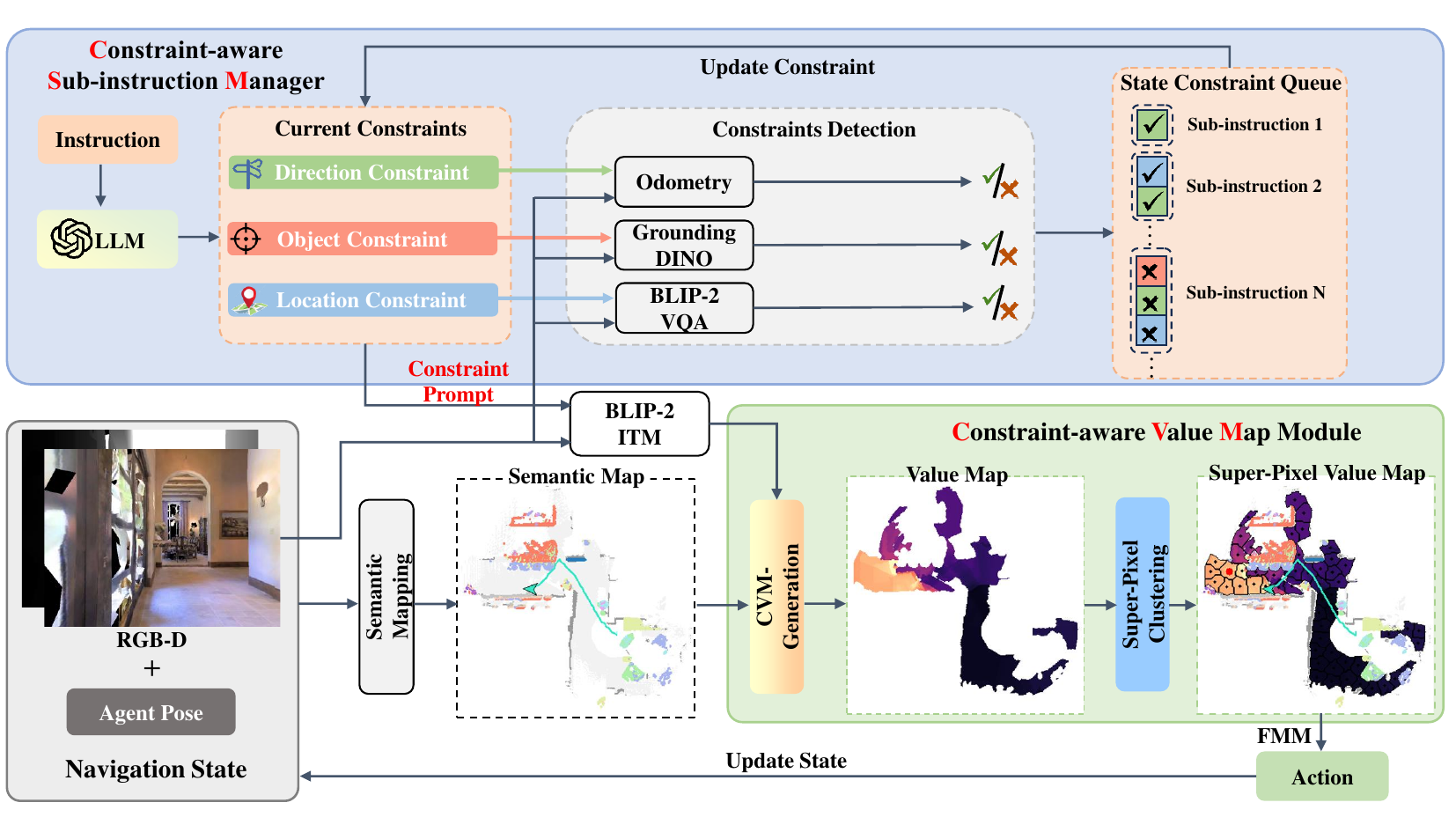}
  \caption{An overall pipeline of CA-Nav. The details of the Constraint-aware Value Map Generation are shown in Figure~\ref{fig:vlmap}.}
  \label{fig:framework}
\end{figure*}

As illustrated in Figure~\ref{fig:framework}, for each episode, CSM and CVM work coherently to execute instruction-following navigation. 
The CSM identifies key constraints that define the completion criteria for each sub-instruction, ensuring constraint-aware sub-instruction switching and navigation progress tracking. 
Upon completing a sub-instruction, CSM uses these constraints to automatically transit to the next one (\S~\ref{SIM}). 
Then the CVM uses the identified constraints to build a constraint-aware value map, 
which guides navigation for the current sub-instruction (\S~\ref{SVM}).

\subsection{Constraint-aware Sub-instruction Manager}
\label{SIM}
The Constraint-aware Sub-instruction Manager (CSM) aims to decompose instructions and track the navigation progress through explicit sub-instruction switching. 
We achieve this by appropriately prompting an LLM and designing a constraint-aware switching mechanism.

\textbf{Instruction Decomposition.} 
\label{instruction decomposition}
As shown in Figure~\ref{fig:teaser}, at the beginning of an episode, CSM decomposes the instruction into a sequence of sub-instructions. 
Each of them outlines the goal for the current sub-instruction and specifies the constraints for switching to the next one.
We categorize these constraints as object constraints (e.g., ``chair"), location constraints (e.g., ``bedroom"), and direction constraints (e.g., ``turn left").
Particularly, object and location constraints describe the landmarks that can be observed along the desired navigation path. 
Thus, they will prompt subsequent building of the value map (\S~\ref{SVM}).
In practice, we implement the above decomposition and constraints extraction process through an LLM.
The prompt consists of four parts, namely task description, output definition, few-shot prompt, and key content reminder (in Appendix Figure 5). 
We find that it's helpful to give large language models an example to follow. 
Because the few-shot prompt can set a clear expectation of the desired output. 

\textbf{Sub-instruction Switching.} 
\label{Sub-instruction Switching}
In the zero-shot VLN-CE setting, the agent should be aware of its navigation progress and automatically switch sub-instructions.
We achieve this through a constraint-aware sub-instruction switching mechanism. 
As shown in Figure~\ref{fig:framework}, CSM maintains a queue in the order of decomposed sub-instructions, with each element containing a set of constraints. 
These constraints can be a combination of object, location, and direction constraints, depending on the current sub-instruction. 
CSM always selects the first unsatisfied constraint set as the current and only switches to the next set once all constraints in the current set are satisfied. 
The agent sequentially checks each constraint within the current set at each step.
We leverage VLM and odometry information to design satisfaction checks for the constraints:
\begin{itemize}[itemsep=2pt,topsep=0pt,parsep=0pt, leftmargin=1em]
\item Object Constraints: Since the navigation instructions include open vocabulary, we use Grounding DINO~\cite{liu2023grounding} to detect objects. A constraint is considered satisfied if the agent detects the object within a certain range $r$ which could empirically be set as 5 meters. 
\item Location Constraints: Indoor location detection can be approached as scene recognition, using a Visual Question Answering (VQA) model such as BLIP2~\cite{li2023blip2} with the template: ``Can you see the \textit{$<location>$}?". If the answer is yes, the constraint is considered satisfied.
\item Direction Constraints: The change in direction should be assessed based on the agent's trajectory rather than just its orientation. To do this, we query the odometry for the poses within a certain time window $\tau$, which could be empirically selected as 5. 
The poses are denoted as $\mathbf{p_{t}}$ and $\mathbf{p_{t-\tau}}$. The change of direction and angle is then calculated using their cross-product and dot product.
\end{itemize}
Sometimes, the agent may get stuck on a single constraint or switch between constraints too frequently. 
To address this, we empirically establish a maximum step threshold of 25 to prompt the agent to switch constraints when progress stalls, along with a minimum step threshold of 10 to ensure adequate focus on each constraint before switching.

\begin{figure*}[t]
  \centering
  \includegraphics[width=0.9\textwidth, trim=0 80 0 70, clip]{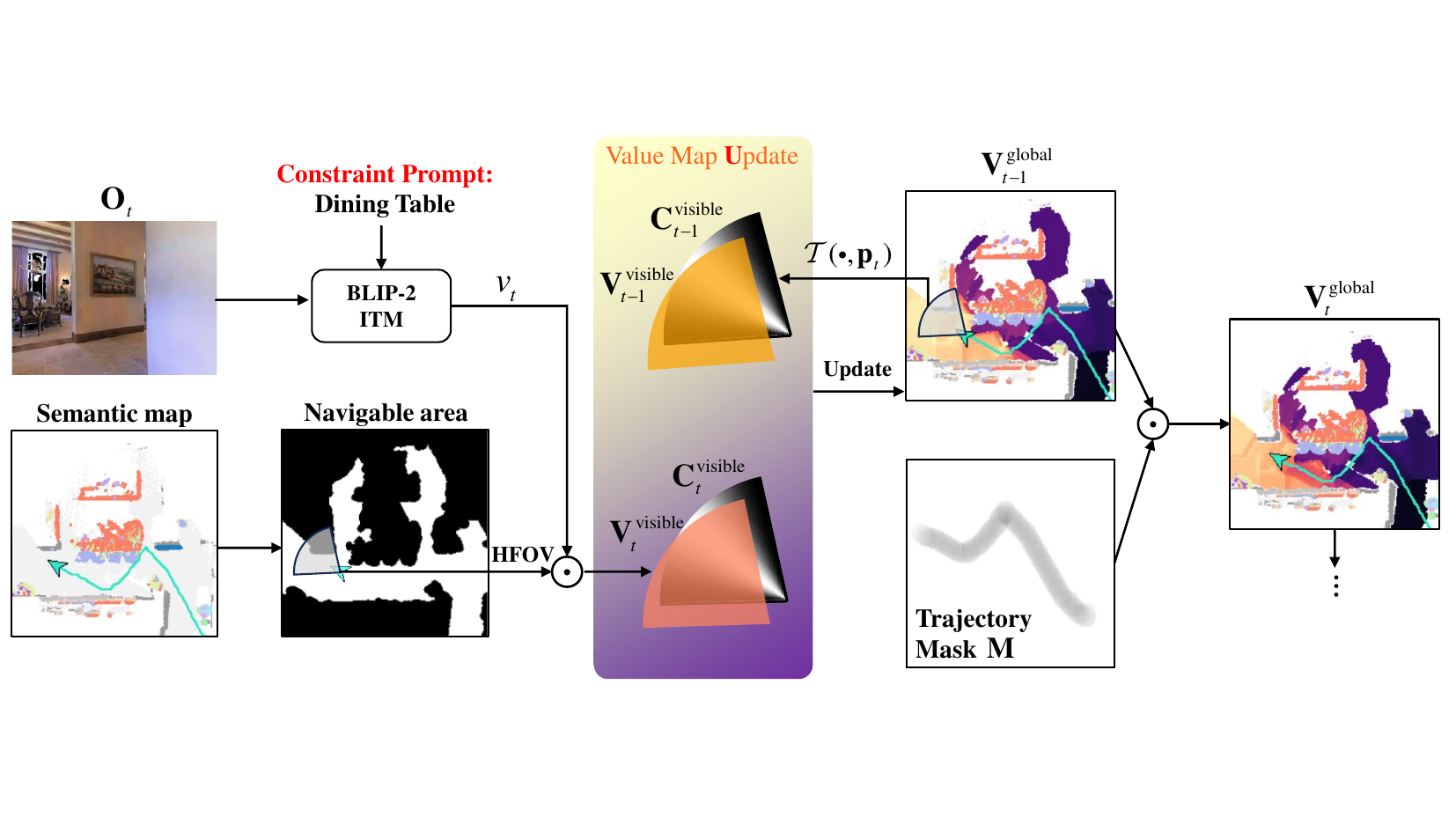}
  \caption{Details of the Constraint-aware Value Map Generation.}
  \label{fig:vlmap}
\end{figure*}
\subsection{Constraint-aware Value Mapper}
\label{SVM}
After CSM provides the current sub-instruction along with its associated constraints, these constraints serve as guiding factors for the value map construction. 
Specifically, we propose a Constraint-aware Value Mapper (CVM) to ground the constraints within the visual environment, ensuring that the value map reflects the potential of each observation to satisfy the current constraints.
The CVM is then refined using a superpixel clustering method for waypoint selection.

\textbf{Constraint-aware Value Map Generation.}
\label{value map generation}
As navigation progresses, we maintain and continuously update a value map that encodes both the environment layout and instruction-relevant values. 
Specifically, the environment layout is captured by maintaining a semantic map, from which we derive navigable areas following SemExp~\cite{chaplot2020semexp}. 
Below, we describe how the instruction-relevant values are projected onto these navigable areas. 
An overview of this process is illustrated in Figure~\ref{fig:vlmap}.
Recall that in our setting the agent uses egocentric observation space so the visible region is limited by its HFOV.
Therefore, the value map $\mathbf{V}_t^{\text{global}}$ can be divided into two parts, namely the invisible part $\mathbf{V}_t^{\text{invisible}}$ and the visible part $\mathbf{V}_t^{\text{visible}}$ where the value update takes place. Formally:
\begin{equation}
\label{eq:global value map update}
\mathbf{V}_t^{\text{global}}=\left[\left(\mathbf{V}_t^{\text{visible}} + \mathbf{V}_t^{\text{invisible}}\right)\cdot \gamma^{\mathcal{B}}\right] \odot \mathbf{M}
\end{equation}
where a decay factor $\gamma$ and a trajectory map $\mathbf{M}$ are include in the global value map update function which will be interpreted in the following parts.

At each time step, given the current RGB observation $\mathbf{O}_t$ and the constraint prompt provided by the Constraint Semantic Module (CSM) (\S~\ref{instruction decomposition}), we first compute the cosine similarity between them.
This similarity score $v_t = \text{BLIP2}(\mathbf{O}_t, \textrm{prompt}_t)$ reflects how relevant the current observation is to the given instruction constraint. 
The similarity score $v_t$ is applied to the corresponding frustum region and forms the $\mathbf{V_t^{\text{visible}}}$.
When updating the value map we follow VLFM~\cite{yokoyama2024vlfm} that utilizes a confidence mask to apply a cosine-weighted average:
\begin{equation}
\mathbf{C}_t = \left\{
\begin{array}{ll}
\text{cos}^2(\theta_{i,j}/(\theta_{\text{hfov}}/2)\cdot\pi/2), &\text{if $(i,j)$ is visible}\\
0, &\text{if $(i,j)$ is invisible}
\end{array}
\right.
\end{equation}
where $\theta_{i,j}$ denotes the angle between pixel $(i,j)$ and the camera's optical axis, and $\theta_{\text{hfov}}$ is the angle of HFOV. 
The confidence mask $\mathbf{C}$ assigns higher weights to regions near the optical axis, prioritizing areas with more reliable observations, while assigning zero weight to regions outside the frustum.

We then update the value map following Equation~\ref{eq: update_visible}, the function $\mathcal{T}$ extracts the frustum region from the previous step of global value map according to current agent pose $\mathbf{p}_t$ and forms $\mathbf{V}_{t-1}^{\text{visible}}$, then the confidence mask in visible regions are used to do the cosine-weighted average update.

\vspace{-6pt}
\begin{equation}
\label{eq: update_visible}
\begin{alignedat}{2}
&\mathbf{V}_t^{\text{visible}} &&\leftarrow \frac{\mathbf{C}_t^{\text{visible}}\mathbf{V}_t^{\text{visible}}+\mathbf{C}_{t-1}^{\text{visible}}\mathbf{V}_{t-1}^{\text{visible}}}{\mathbf{C}_t^{\text{visible}} + \mathbf{C}_{t-1}^{\text{visible}}}\\
&\mathbf{V}_{t-1}^{\text{visible}} &&\leftarrow \mathcal{T}\left(\mathbf{V}_{t-1}^{\text{global}}, \mathbf{p}_t\right)\\
&\mathbf{V}_t^{\text{invisible}} &&\leftarrow \mathbf{V}_{t-1}^{\text{invisible}}; \mathbf{V}_t^{\text{invisible}} \text{ is initialized as \textbf{0}}\\
\end{alignedat}
\end{equation}
\vspace{-6pt}

After updating the value map we update the confidence mask following Equation~\ref{eq: update_confidence} which is similar to the process of updating value map.
\vspace{-6pt}
\begin{equation}
\label{eq: update_confidence}
\begin{alignedat}{2}
&\mathbf{C}_t^{\text{visible}}&&\leftarrow\frac{\left(\mathbf{C}_t^{\text{visible}}\right)^2+\left(\mathbf{C}_{t-1}^{\text{visible}}\right)^2}{\mathbf{C}_t^{\text{visible}} + \mathbf{C}_{t-1}^{\text{visible}}}\\
&\mathbf{C}_{t-1}^{\text{visible}} &&\leftarrow \mathcal{T}\left(\mathbf{C}_{t-1}^{\text{global}}, \mathbf{p}_t\right)\\
&\mathbf{C}_t^{\text{invisible}} &&\leftarrow \mathbf{C}_{t-1}^{\text{invisible}}; \mathbf{C}_t^{\text{invisible}} \text{ is initialized as \textbf{0}}
\end{alignedat}
\end{equation}
\vspace{-6pt}

It's worth noting that when CSM switches constraints, the constraint prompt changes.
If the previous value map is cleared entirely, the agent loses navigational cues and must rebuild from scratch, often causing it to linger and collide.
To address this, we introduce a historical decay factor $\gamma$ that retains past value map with a reduced weight, helping the agent focus on new constraints while still leveraging past exploration.
The historical decay factor is controlled by a binary indicator $\mathcal{B}$ that reflects whether a constraint switch occurs.
The decay of the value map happens only when constraint switches.
The default $\gamma$ is set to 0.5.

\vspace{-6pt}
\begin{equation}
\gamma \in (0, 1];\mathcal{B} = \left\{
\begin{array}{ll}
1, \text{switch constraint}\\
0, \text{not switch constraint}
\end{array}
\right.
\end{equation}
\vspace{-1em}

Finally, to encourage exploration, we introduce a trajectory mask $\mathbf{M}$ representing the agent's willingness to explore, as shown in Equation~\ref{eq: trajectory_mask}. 
This mask starts as a matrix of ones, but its values decay exponentially by a factor of $\lambda$ in regions the agent has already traversed, $k(i,j)$ represents the total number of visits to position $(i,j)$.
The value map will adjusted by element-wise multiplication with this trajectory mas.
By default, the $\lambda$ is set to 0.95.
\vspace{-2pt}
\begin{equation}
\label{eq: trajectory_mask}
\mathbf{M}_{i,j} = \lambda^{k(i,j)}, \lambda \in (0,1],\mathbf{M} \text{ is initialized as \textbf{1}}
\end{equation}

\textbf{Superpixel-based Waypoint Selection.} 
\label{Superpixel-based Waypoint Selection}
Next, we need to choose a waypoint according to the value map. A common approach is using frontier-based exploration (FBE)~\cite{yamauchi1997frontier} which has been widely used in object navigation task~\cite{yokoyama2024vlfm, zhou2023esc, shah2023navigation}.
Among them, VLFM~\cite{yokoyama2024vlfm} is the most comparable to ours, as it also constructs a value map and selects the frontier with the highest value as the navigation target. 
However, it only focuses on the boundaries of the explored area, restricting the full utilization of the value map. 
We propose that using the full value map can benefit navigation, however, our CSM introduces sub-instruction switching, which may lead to abrupt value changes at the frontiers.
To that end, we propose a superpixel-based waypoint selection approach that considers the global value map.

Specifically, we employ SLIC~\cite{achanta2012slic} to refine the constraint-aware value map. 
Given current value map $\mathbf{V}$, SLIC produces a set of superpixels $\{ \mathbf{S_1}, \mathbf{S_2}, \cdots, \mathbf{S_n} \}$, where each superpixel $\mathbf{S_i}$ represents a visually consistent region. 
Let $v(p)$ denote the value at pixel $p$ in $\mathbf{V}$, then compute the average value of each superpixel $\mathbf{S_i}$. 
The optimal region $\mathbf{S^*}$ is then selected based on the highest average value, guiding the agent towards areas of greater semantic relevance:
{\small
\begin{equation}
\label{eq:action_fuse}
\begin{aligned}
\mathbf{V(S_i)} &= \frac{1}{|\mathbf{S_i}|}\sum_{p \in \mathbf{S_i}}v(p),\\
\mathbf{S^*} &= \mathop{\arg\max}\limits_{\mathbf{S_i}}\mathbf{V(S_i)}
\end{aligned}
\end{equation}
}%
Finally, the waypoint is the geometric center of the optimal region $\mathbf{S^*}$.
Note that when the agent reaches the final sub-instruction, it extracts the target's segmentation mask using RepViT-SAM~\cite{wang2023repvitsam, wang2024repvit}, which achieves real-time segmentation of anything.
The mask is then projected onto the semantic map and its geometric center will be the destination waypoint.
After determining the waypoint, the Fast Marching Method (FMM)~\cite{sethian1999fast} is used to plan low-level actions to the waypoint.

\section{Experiments And Results}
\label{experiments}
\subsection{Experiment Setup}
\label{experiment setup}
\textbf{Dataset and Evaluation.}\space 
We primarily conduct experiments on the val-unseen split of R2R-CE~\cite{krantz_vlnce_2020} and RxR-CE~\cite{ku2020room} using Habitat simulator~\cite{savva2019habitat}, which are the only two VLN-CE datasets.
R2R-CE and RxR-CE datasets provide 1839 and 3669 step-by-step trajectory-instruction pairs across 11 unseen environments, respectively, with RxR-CE containing more detailed instructions and longer trajectories.

We use standard metrics following~\cite{anderson2018vision, krantz_vlnce_2020}: Navigation Error (NE) measures the mean distance from the final location to the destination; Success Rate (SR) represents the proportion of episodes with NE under 3 meters; Oracle Success Rate (OSR) calculates SR with an oracle stop policy; Success weighted by Trajectory Length (SPL) normalizes SR by trajectory length; Normalized Dynamic Time Warping (NDTW) evaluates the fidelity between the agent's and the annotated trajectories; and Success-weighted Dynamic Time Warping (SDTW) weights NDTW by SR. For R2R-CE dataset, SR and SPL are the primary metrics to evaluate navigation performance. RxR-CE additionally emphasizes NDTW and SDTW metrics to assess path fidelity.

\vspace{-6pt}
\subsection{Main Results}
\textbf{Methods for Comparison.} We compare CA-Nav with VLN-CE methods that also use low-level actions. 
These methods can be categorized into two types: the first type is training-based methods, including Sara~\cite{irshad2022semantically}, Seq2Seq~\cite{krantz_vlnce_2020}, \vlnbert~\cite{hong_2022_bridging_the_gap}, AG-CMTP~\cite{chen2021topological}, WS-MGMap~\cite{chen2022weakly} and LAW~\cite{raychaudhuri2021law}.
The second type is zero-shot methods, including CoW~\cite{gadre2023cows}, NavGPT-CE~\cite{zhou2024navgpt}, OpenNav~\cite{openai2024gpt4}, InstructNav~\cite{long2024instnav}, A$^2$Nav~\cite{chen2023A2Nav} and our CA-Nav. 
CoW was originally designed for zero-shot object navigation~\cite{majumdar2022zson, zhou2023esc, shah2023navigation}. We adapt it to the VLN-CE task by extracting object and location constraints from the instructions and have CoW navigate to them sequentially. 
We evaluate two representative variants: CoW-OWL and CoW-CLIP-Grad, which use object detectors and gradient-based saliency maps to localize target locations, respectively.

NavGPT was originally for zero-shot VLN, we transfer it to zero-shot VLN-CE by using a waypoint predictor~\cite{hong_2022_bridging_the_gap} to predict navigable nodes in panoramas, discretizing the continuous environment for high-level navigation planning.
We also list other methods that rely on high-level actions, including Sim2Sim~\cite{krantz2022sim2sim}, GridMM~\cite{wang2023gridmm}, ETPNav~\cite{an2024etpnav}, and BEVBert~\cite{an2023bevbert}.

\begin{table*}[ht]
\centering
\small
\caption{Comparison with SOTA methods on R2R-CE validation-unseen split. 
In the Efficient LLM Usage column: \textbf{-} means the LLM is not used, \ding{51} means the LLM is only used before navigation starts, and \ding{55} means the LLM is accessed for each navigation decision.
$\dag$: Our reproduced NavGPT-CE for VLN-CE. $*$: Methods use the same waypoint predictor proposed in~\cite{hong_2022_bridging_the_gap}. \halfcheck: A$^{2}$Nav pretrained the navigator on an action-specific dataset built from HM3D~\cite{hm3d}}.
\resizebox{\linewidth}{!}{
	{\renewcommand{\arraystretch}{1}
\begin{tabular}{p{3.5cm} >{\centering\arraybackslash\columncolor{Gray}}p{1.5cm} >{\centering\arraybackslash\columncolor{Gray}}p{3cm} >{\centering\arraybackslash\columncolor{Gray}}p{2cm} >{\centering\arraybackslash}p{1.2cm} >
{\centering\arraybackslash}p{1.2cm} >
{\centering\arraybackslash}p{1.2cm} >
{\centering\arraybackslash}p{1.2cm}}
\toprule
Method & Zero-shot & Efficient LLM Usage & Egocentric Obs & NE$\downarrow$ & OSR$\uparrow$  & \textbf{SR}$\uparrow$ & \textbf{SPL}$\uparrow$\\
\midrule
Sasra~\cite{irshad2022semantically}  & \ding{55} & \textbf{-} & \ding{55} & 8.32 & \textbf{-} & 24.0 & 22.0\\
Seq2Seq~\cite{krantz_vlnce_2020} & \ding{55} & \textbf{-} & \ding{55} & 7.77 & 37.0 & 25.0 & 22.0 \\
AG-CMTP~\cite{chen2021topological} & \ding{55} & \textbf{-} & \ding{55} & 7.90 & 39.2 & 23.1 & 19.1\\
\vlnbert~\cite{hong_2022_bridging_the_gap} & \ding{55} & \textbf{-} & \ding{55} & 7.66 & \textbf{-} & 23.2 & 21.7\\ 
\vlnbert$^{*}$~\cite{hong_2022_bridging_the_gap} & \ding{55} & \textbf{-} & \ding{55} & 5.74 & 53.0 & 44.0 & 39.0\\ 
Sim2Sim$^{*}$~\cite{krantz2022sim2sim} & \ding{55} & \textbf{-} & \ding{55} & 6.07 & 52.0 & 43.0 & 36.0\\
GridMM$^{*}$~\cite{wang2023gridmm} & \ding{55} & \textbf{-} & \ding{55} & 5.11 & 61.0 & 49.0 & 41.0 \\
ETPNav$^{*}$~\cite{an2024etpnav} & \ding{55} & \textbf{-} & \ding{55} & 4.71 & 65.0 & 57.0 & 49.0\\
BEVBert$^{*}$~\cite{an2023bevbert} & \ding{55} & \textbf{-} & \ding{55} & 4.57 & 67.0 & 59.0 & 50.0\\
WS-MGMap~\cite{chen2022weakly} & \ding{55} & \textbf{-} & \ding{51} & 6.28 & 47.6 & 38.9 & 34.3\\
\midrule
CoW-OWL~\cite{gadre2023cows} & \ding{51} & \textbf{-} & \ding{51} & 8.72 & 5.9 & 3.4 & 1.6 \\
CoW-CLIP-Grad~\cite{gadre2023cows} & \ding{51} & \textbf{-} & \ding{51} & 8.68 & 3.4 & 1.8 & 1.3 \\
NavGPT-CE$^{* \dag}$~\cite{zhou2024navgpt} & \ding{51} & \ding{55} & \ding{55} & 8.37 & 26.9 & 16.3 & 10.2\\
OpenNav$^*$ (LLama3.1)~\cite{qiao2024opennav} & \ding{51} & \ding{55} & \ding{55} & \textbf{7.25} & 23.0 & 16.0 & \textbf{12.9}\\
InstructNav (ego)~\cite{long2024instnav} & \ding{51} & \ding{55} & \ding{51} & 9.20 & 47.0 & 17.0 & 11.0\\
A$^{2}$Nav~\cite{chen2023A2Nav} & \halfcheck & \ding{51} & \ding{51} & \textbf{-} & \textbf{-} & 22.6 & 11.1\\
\textbf{CA-Nav} & \ding{51} & \ding{51} & \ding{51} & 7.58 & \textbf{48.0} & \textbf{25.3} & 10.8\\ 
CA-Nav (test-unseen) & \ding{51} & \ding{51} & \ding{51} & 8.32 & 39.4 & 23.0 & 11.0\\ 
\bottomrule
\end{tabular}}}
\label{sota_r2r}
\end{table*}
\textbf{R2R-CE Dataset.}
\label{r2r-ce}
As shown in Table \ref{sota_r2r}, our method surpasses various models trained with panorama in NE, SR, and OSR. 
This indicates that building an egocentric zero-shot VLN-CE system with LLMs and VLMs is feasible and potential.
Our method's success with egocentric observations lies in the value map's dual role: storing environmental layout memory and leveraging VLM's prior knowledge.

Both the detector-based and gradient-based variants of CoW perform poorly, which is unsurprising. 
The main reason is that CoW's single-object searching capability is inherently limited\footnote{CoW-CLIP-Grad achieved 9.2 SR and CoW-OWL achieved 4.7 SR in Habitat MP3D.}.
The performance further degrades in VLN-CE, where the task involves multi-object search and is subject to accumulating navigation errors.
Moreover, CoW is designed for label-level object navigation (i.e., finding any instance of a given object category), while VLN-CE requires instance-level understanding - the agent needs to locate specific objects or locations that match detailed language descriptions, rather than just finding any object of a certain category.

To study the performance of zero-shot methods in continuous environments, we transfer NavGPT\footnote{NavGPT achieved 34.0 SR and 29.0 SPL on the R2R dataset.} to continuous environments (i.e., NavGPT-CE), with results showing a success rate drop of over 50$\%$. 
One potential issue is that the caption model sometimes misses crucial details like landmarks and spatial layout. 
Additionally, the LLM may generate hallucinations when summarizing navigation history or reasoning about the agent's status, resulting in incorrect navigation planning. (visualization analysis in \S~\ref{navigation visualization}).

Similar to NavGPT-CE, OpenNav also utilizes panoramic observations and a waypoint predictor~\cite{hong_2022_bridging_the_gap} to generate candidate waypoints in continuous environments. 
It then leverages open-source VLMs to describe the observations and uses CoT reasoning to make navigation decisions based on open-source LLMs. 
The primary reasons for this are similar to the drawbacks of NavGPT-CE, namely, that the descriptions often overlook significant visual details, despite OpenNav's consideration of spatial information.
 
Similar to our approach, InstructNav also explores the potential of value maps.
At each decision step, it invokes GPT-4 to plan the next navigation sub-task, then prompts GPT-4V with the current panoramic observations to estimate the most appropriate heading. Additionally, it constructs other multi-sourced value maps using hand-crafted rules to guide waypoint selection.
The use of panoramic observations, along with the powerful GPT-4 and GPT-4V, contributes to InstructNav's high performance which achieves 31.0 SR.
However, this distinguishes it from our task setting, where CA-Nav utilizes egocentric views and calls LLMs more efficiently.
When InstructNav also use egocentric observations the performance drops significantly (from 31.0 SR to 17.0 SR).

The method closest to our task setting is A$^{2}$Nav, which also uses LLM efficiently and utilizes egocentric observation. 
Our method surpasses it on SR and OSR because A$^{2}$Nav focuses more on actions. 
Descriptions of actions in instructions are more prone to ambiguity. 
For example, a path to the living room might be described as either ``Turn left to the living room." or ``Turn slightly right, then turn left immediately and go to the living room.". 
We conclude that rigid execution of actions such as turning tends to cause serious cumulative errors, and the agent should focus on more clearly described instructions such as landmarks.

However, CA-Nav exhibits a slightly lower SPL compared to some other methods. 
This is primarily because, in the egocentric setting, the agent receives more limited observations compared to panoramic views. 
As a result, the agent spends more steps exploring the environment, which slightly reduces path efficiency. 
In contrast, OpenNav utilizes panoramic observations along with a waypoint predictor, providing additional information and thereby achieving a higher SPL. 
Similarly, InstructNav achieves an SPL of 24 with panoramic views, but it drops to 11 when using egocentric observations. 
Therefore, under the egocentric observation setting, CA-Nav achieves a comparable SPL to other methods.
We also report CA-Nav's performance on the test-unseen split\footnote{To our knowledge, no other zero-shot VLN-CE methods have reported results on the test unseen split.}, which demonstrates a comparable SR and a slightly higher SPL compared to its performance on the val-unseen split. 
This indicates that CA-Nav maintains its navigation effectiveness in previously unseen environments. 
Moreover, the improvement in SPL also highlights its ability to generalize to new instructions and adapt to unfamiliar spatial layouts.

Overall, CA-Nav achieves state-of-the-art performance in SR, NE, and OSR under the setting of using egocentric observations and performs comparably on SPL.

\begin{table*}[ht]
\centering
\caption{Comparison with SOTA methods on RxR-Habitat val-unseen split (only English).}
\resizebox{\linewidth}{!}{
{\renewcommand{\arraystretch}{0.8}
\begin{tabular}{p{2.4cm} >{\centering\arraybackslash\columncolor{Gray}}p{1.3cm} >{\centering\arraybackslash\columncolor{Gray}}p{2.5cm} >{\centering\arraybackslash\columncolor{Gray}}p{1.8cm} >{\centering\arraybackslash}p{0.8cm} >
{\centering\arraybackslash}p{0.8cm} >
{\centering\arraybackslash}p{0.8cm} >
{\centering\arraybackslash}p{0.8cm} >
{\centering\arraybackslash}p{1.cm}}
\toprule
Method & Zero-shot & Efficient LLM Usage & Egocentric Obs & NE$\downarrow$ & \textbf{SR}$\uparrow$ & \textbf{SPL}$\uparrow$ & \textbf{NDTW}$\uparrow$ & \textbf{SDTW}$\uparrow$\\
\midrule
LAW~\cite{raychaudhuri2021law}  & \ding{55} & \textbf{-} & \ding{55} & 11.04 & 10.0 & 9.0 & 37.0 & 8.0\\
\vlnbert$^{*}$~\cite{hong_2022_bridging_the_gap} & \ding{55} & \textbf{-} & \ding{55} & 8.98 & 27.1 & 23.7 & 46.7 & \textbf{-}\\ 
GridMM$^{*}$~\cite{wang2023gridmm} & \ding{55} & \textbf{-} & \ding{55} & 8.42 & 36.3 & 30.1 & 48.2 & 33.7\\
ETPNav$^{*}$~\cite{an2024etpnav} & \ding{55} & \textbf{-} & \ding{55} & 5.64 & 54.8 & 44.9 & 61.9 & 45.3\\
WS-MGMap~\cite{chen2022weakly} & \ding{55} & \textbf{-} & \ding{51} & 9.83 & 15.0 & 12.1 & \textbf{-} & \textbf{-}\\
\midrule
A$^{2}$Nav~\cite{chen2023A2Nav} & \halfcheck & \ding{51} & \ding{51} & \textbf{-} & 16.8 & \textbf{6.3} & \textbf{-} & \textbf{-}\\
\textbf{CA-Nav} & \ding{51} & \ding{51} & \ding{51} & \textbf{10.37} & \textbf{19.0} & 6.0 & \textbf{13.5} & \textbf{5.0} \\ 
\bottomrule
\end{tabular}}
}
\label{sota_rxr}
\end{table*}
\textbf{RxR-CE Dataset.} 
\label{rxr-ce}
Table \ref{sota_rxr} reports the results on the RxR-CE val-unseen split\footnote{The RxR-CE test-unseen \href{https://eval.ai/web/challenges/challenge-page/719/leaderboard/1966/success}{leaderboard} has been closed, and the corresponding results are currently unavailable.}, which contains longer and more detailed instructions with fine-grained descriptions of landmarks and actions. 
These characteristics of the dataset make it more challenging, as they lead to more frequent instruction switches and more complex constraint identification. 
Despite these challenges, CA-Nav demonstrates competitive performance, surpassing several models trained with panoramic observations such as LAW or egocentric observations such as WS-MGMap.
Notably, CA-Nav outperforms A$^{2}$Nav in both SR and NE, while remaining competitive in SPL. 
Although A$^{2}$Nav pretrained five action-aware navigators it shows lower performance in SR and NE, indicating that CA-Nav's CSM and CSV are effective even when facing complex instructions.

\vspace{-3pt}
\subsection{Ablation Study}
We conduct ablation experiments on the R2R-CE dataset to evaluate each component of CA-Nav, focusing on six aspects: (1) the impact of constraints, (2) value map update methods, (3) the effectiveness of superpixel-based waypoint selection, (4) the impact of different superpixel size, (5) robustness to different constraint switching step thresholds, (6) analysis of different history decay factors, (7) influence of trajectory decay factors and (8) generalization to different LLMs.

\textbf{The effect of different constraints.}
We begin our investigation by examining how the performance of CA-Nav is influenced by different types of constraints. 
As shown in Table~\ref{ablations}-1, we first remove all constraints except those related to the final sub-instruction, making the task more akin to object navigation.
The success rate drops by about 21$\%$ compared to our best performance.
Next, we evaluate the performance of using each constraint independently.
The results show notable improvements, particularly when employing object constraints.
We then explore the effect of combining two types of constraints.  
However, performance declines when direction constraints are included (Row 5 and Row 6), especially in terms of SPL. 
The reason is that relying solely on direction constraints while lacking object or location constraints leads to excessive exploration, resulting in lower SPL.
Notably, combining object and location constraints leads to significant improvements (Row 7), indicating that these two factors contribute the most to effective navigation.
Only when both object and location constraints are incorporated does the direction constraint become beneficial, helping the agent better follow sub-instructions and ultimately achieve the best performance.

This further supports our analysis in \S~\ref{r2r-ce} that the agent relies more on landmarks than directions during navigation, which also explains why CA-Nav outperforms A$^{2}$Nav.
Finally, when both object.
By designing appropriate constraints and using a Constraint-aware Sub-instruction Manager, our method can autonomously switch between sub-instructions in long-horizon navigation tasks.
This not only compensates for VLFM's~\cite{yokoyama2024vlfm} inability to handle sequential tasks but also avoids frequent calls to LLMs.

\textbf{How can the value map be better updated?} 
Recall that in the value map generation section \S~\ref{value map generation}, we designed a historical decay mechanism to leverage past explorations and a trajectory mask to encourage exploration. 
We investigate their influence in Table~\ref{ablations}-3.
In Row1, we ablate both historical decay and trajectory mask which means the value map will be completely reset upon constraint switching. 
It's not unexpected that this brings a severe performance drop. 
In Row 2, we ablate the trajectory mask, the performance drops and we observe that the agent tends to get stuck in narrow spaces such as corridors.
This suggests that the trajectory mask contributes positively to the agent's exploration.
In Row 3, where the historical decay is ablated and shows a worse performance. 
This highlights the inefficiency of discarding all previous knowledge after each constraint switches, as it forces the agent to rediscover previously explored areas.
However, Row 4, which incorporates both a trajectory mask and long-term value maintenance using historical decay, effectively balances exploration with exploitation and achieves the best performance.

\begin{table*}[t]
\centering
\caption{Ablation results for constraints, switching step thresholds, value map update methods, waypoint selection methods, history decay factors, trajectory decay factors, superpixel sizes and LLMs.}
\fontsize{8}{8}\selectfont
\label{ablations}
\resizebox{\linewidth}{!}{
	{\renewcommand{\arraystretch}{1.2}
\begin{tabular}{ccccccc|ccccc}
\hline
\rowcolor{gray!30}\multicolumn{7}{c}{1. The effect of different constraints} & \multicolumn{5}{c}{2. Different constraint 
 switching step thresholds}\\
\hline\hline
Direction & Object & Location & NE$\downarrow$ & OSR$\uparrow$ & \textbf{SR}$\uparrow$ & \textbf{SPL}$\uparrow$ & Min/Max & NE$\downarrow$ & OSR$\uparrow$ & \textbf{SR}$\uparrow$ & \textbf{SPL}$\uparrow$ \\
\hline
\textbf{-} &\textbf{-} &\textbf{-} & 7.95 & 36.4 & 20.0 & 9.8 & - & - & - & - & - \\
\ding{51} &\textbf{-} &\textbf{-} & 7.79 & 40.4 & 22.8 & 10.1 & 0/0 & 7.81 & 40.2 & 22.3 & 10.9 \\\cline{8-12}
\textbf{-}& \ding{51} & \textbf{-}& 7.71 & 44.0 & 23.8 & 10.7 & 5/25 & 7.73 & 45.1 & 24.1 & 10.9 \\
\textbf{-}& \textbf{-}& \ding{51} & 7.85 & 43.6 & 23.5 & 10.4 & 10/25 & \textbf{7.58} & \textbf{48.0} & \textbf{25.3} & \textbf{10.8} \\
\ding{51}& \textbf{-}& \ding{51}& 8.10& 45.4& 20.9& 7.2 & 15/25 & 7.65 & 47.6 & 24.8 & 9.9 \\\cline{8-12}
\ding{51}& \ding{51}&\textbf{-} & 7.93& 46.6& 23.1& 8.6  & 10/15 & 7.54 & 48.1 & 25.0 & 11.7 \\
\textbf{-}& \ding{51} & \ding{51}& 7.74& 46.9& 24.0& \textbf{10.9} & 10/25 & \textbf{7.58} & \textbf{48.0} & \textbf{25.3} & \textbf{10.8} \\
\ding{51} & \ding{51} & \ding{51} & \textbf{7.58}& \textbf{48.0}& \textbf{25.3}& 10.8 & 10/35 & 7.56 & 48.5 & 24.4 & 10.2 \\
\hline\hline
\rowcolor{gray!30}\multicolumn{7}{c}{3. Influence of value map update methods} & \multicolumn{5}{c}{4. Different waypoint selection methods}\\
\hline
\multicolumn{3}{c}{Method} & NE$\downarrow$ & OSR$\uparrow$ & \textbf{SR}$\uparrow$ & \textbf{SPL}$\uparrow$ & Method & NE$\downarrow$ & OSR$\uparrow$ & \textbf{SR}$\uparrow$ & \textbf{SPL}$\uparrow$\\
\hline
\multicolumn{3}{c}{None} & 7.68 & 38.9 & 22.3 & 10.3 & FBE-based & 8.08 & \textbf{50.2} & 21.9 & 10.4\\
\multicolumn{3}{c}{trajectory mask} & 7.65 & 41.3 & 24.6 & \textbf{10.9} & Pixel-based & 7.87 & 42.9 & 22.9 & 10.4\\
\multicolumn{3}{c}{historical decay} & \textbf{7.57} & 45.6 & 24.6 & 10.8 & {ORP-based} & \textbf{7.54} & 45.7 & 25.0 & 10.6\\
\multicolumn{3}{c}{trajectory mask + historical decay} & 7.58 & \textbf{48.0} & \textbf{25.3} & 10.8 & Superpixel-based & 7.58 & 48.0 & \textbf{25.3} & \textbf{10.8}\\
\hline\hline
\rowcolor{gray!30}\multicolumn{7}{c}{5. Analysis of different history decay factors} & \multicolumn{5}{c}{6. Influence of trajectory decay factors}\\
\hline
\multicolumn{3}{c}{History factor $\gamma$} & NE$\downarrow$ & OSR$\uparrow$ & \textbf{SR}$\uparrow$ & \textbf{SPL}$\uparrow$ & Trajectory factor $\lambda$ & NE$\downarrow$ & OSR$\uparrow$ & \textbf{SR}$\uparrow$ & \textbf{SPL}$\uparrow$\\
\multicolumn{3}{c}{0.0} & \textbf{7.57} & 45.5 & 24.4 & 10.7 & 0.95 & 7.58 & 48.0 & \textbf{25.3} & \textbf{10.8} \\
\multicolumn{3}{c}{0.25} & 7.93 & 47.2 & 24.8 & 10.7 & 0.90 & \textbf{7.52} & 47.9 & 24.5 & 10.8\\
\multicolumn{3}{c}{0.50} & 7.58 & \textbf{48.0} & \textbf{25.3} & \textbf{10.8} & 0.85 & 7.57 & \textbf{48.8} & 24.7 & 10.8\\
\multicolumn{3}{c}{0.75} & 7.77 & 46.0 & 24.0 & 10.5 & 0.80 & 7.82 & 46.8 & 24.8 & 10.2\\
\hline\hline
\rowcolor{gray!30}\multicolumn{7}{c}{7. Different superpixel sizes} & \multicolumn{5}{c}{8. Generalization to different LLMs}\\
\hline
\multicolumn{3}{c}{Supepixe Size} & NE$\downarrow$ & OSR$\uparrow$ & \textbf{SR}$\uparrow$ & \textbf{SPL}$\uparrow$ & Method & NE$\downarrow$ & OSR$\uparrow$ & \textbf{SR}$\uparrow$ & \textbf{SPL}$\uparrow$\\
\hline
\multicolumn{3}{c}{$25 \times 25$ (pixel)} & \textbf{7.54} & \textbf{47.5} & 23.7 & 10.3 & GPT-3.5 & 7.66 & 45.0 & 21.1 & 9.4\\
\multicolumn{3}{c}{$50 \times 50$ (pixel)} & 7.57 & 46.5 & \textbf{24.9} & \textbf{10.5} & Claude-3.5 Sonnet & \textbf{7.41} & 47.1 & 25.2 & \textbf{11.8}\\
\multicolumn{3}{c}{$75 \times 75$ (pixel)} & 7.75 & 46.0 & 24.1 & 10.1 & GPT-4 & 7.58 & \textbf{48.0} & \textbf{25.3} & 10.8\\
\multicolumn{3}{c}{$100 \times 100$ (pixel)} & 7.89 & 42.7 & 21.4 & 8.7 & - & - & - & - & -\\
\hline
    \end{tabular}}}
\end{table*}
\textbf{Comparison of different waypoint selection methods.} 
\label{waypoint selection methods}
To verify the effectiveness of the superpixel-based waypoint selection method, we compare it with the FBE-based method mentioned in \S~\ref{Superpixel-based Waypoint Selection}.
Additionally, we also compare it to the Pixel-based and ORP-based methods.
The Pixel-based method directly selects the pixel with the highest value from the global value map as the navigation target.
While the ORP-based method chooses the \textbf{O}ptimal \textbf{R}egion first and then selects the \textbf{P}ixel with the highest value as the waypoint.

Results in Table~\ref{ablations}-4 show that the superpixel-based method outperforms the FBE-based and Pixel-based methods in terms of SR, and also takes the lead in NE and SPL. 
This is because value map updates tend to be uneven, making the FBE-based method, which focuses on local frontiers more susceptible to disruptions from sudden value changes (\S~\ref{navigation visualization}). 
While a confidence mask is applied to perform cosine-weighted averaging during value map generation (\S~\ref{value map generation}), the trajectory mask affects the map's smoothness, and the historical decay further disrupts consistency when sub-instructions switch.
This indicates the importance of incorporating a more global perspective into the value map utilizing. 
The ORP-based method performs closely with the superpixel-based method and is superior to FBE-based and Pixel-based methods.
This indicates that clustering similar regions would improve the understanding of the explored area, resulting in more stable navigation planning. 

\textbf{Robustness analysis of superpixel size.}
\label{robustness of superpixel size}
The superpixel size determines the granularity of the waypoints used for navigation. 
Smaller superpixels result in more precise waypoints, while larger superpixels result in coarser regions, which may affect the agent's ability to make accurate navigation decisions.
Therefore, it's necessary to analyze the impact of different superpixel sizes on CA-Nav's performance.
We test four different superpixel sizes and the resolution ratio was set to 5, which means each pixel corresponds to a 5 cm $\times$ 5 cm area in the real world.
The results in Table~\ref{ablations}-7 show that a smaller superpixel size (25 $\times$ 25) leads to lower navigation error (NE) and higher object success rate (OSR). 
This is likely due to the finer granularity of waypoints, which help the agent get closer to the destination. 
However, the success rate (SR) decreases because smaller superpixels are more sensitive to noise in the value map. 
On the other hand, when the superpixel size is too large (e.g., 100 $\times$ 100), the regions become overly coarse, making it difficult for CA-Nav to select precise waypoints, resulting in a significant drop in performance. 
In such cases, the robot may end up in imprecise areas or fail to navigate effectively, particularly when encountering smaller obstacles or narrow paths.
Our results suggest that superpixel sizes around 50 $\times$ 50 achieve an optimal balance between precision and robustness to value map noise.
We finally set the default superpixel size as 48 $\times$ 48 which achieves the best result.

\textbf{Influence of constraint switching step threshold.} 
Recall that in the sub-instruction switching section\S~\ref{Sub-instruction Switching} we introduced minimum and maximum constraint switching step thresholds respectively.
This mechanism is designed to ensure that transitions between constraints occur neither too early nor too late, which could negatively affect navigation performance. 
By setting appropriate thresholds for the minimum and maximum steps, we can improve the stability of navigation.
To figure out its impact on navigation we test various combinations of minimum and maximum thresholds as shown in table~\ref{ablations}-2.
As shown in Line 1, When both the minimum and maximum thresholds are set to 0, representing the vanilla constraints transition mechanism, the SR, OSR, and NE all get worse. 
This demonstrates the importance of the threshold in improving performance.
However, CA-Nav shows robustness against minor variations in other thresholds. 
Whether the minimum is set to 10 or 15, and the maximum to 25 or 35, the performance remains within a narrow range.
This suggests that CA-Nav is not overly sensitive to small changes in the constraint switching steps, making it more flexible and adaptable in real-world scenarios.

\textbf{Analysis of different history decay factor.} 
\label{effect of history factor}
We investigate the impact of the historical decay factor $\gamma$ on final performance in Table~\ref{ablations}-5.
The default setting of $\gamma=0.5$ means that when switching constraints, the historical value map is decayed by half.
Both decreasing and increasing $\gamma$ lead to a drop in SR, though the changes are moderate, and the SPL remains stable.
Overall, a medium decay factor (e.g., 0.5) proves to be optimal, striking a balance between preserving historical information and adapting to new constraints.

\textbf{Influence of trajectory decay factor.} 
\label{effect of trajectory factor}
The influence of trajectory decay factor is also examined, as shown in Table~\ref{ablations}-6.
Specifically, the decay factor $\lambda$ is applied to the trajectory mask to encourage exploration, and it is set to 0.95 by default.
We evaluate the performance of CA-Nav under different $\lambda$ values.
As $\lambda$ decreases, both SR and OSR exhibit a moderate decline, indicating that CA-Nav remains robust to minor variations in $\lambda$.

\textbf{Generalization to other LLMs.} 
\label{effect of LLMs}
We further replace GPT-4 in our method with GPT-3.5 and Claude 3.5 Sonnet to explore the robustness and generalizability across different LLMs. 
From Table \ref{ablations}-8, we observe that Claude-3.5 Sonnet achieves a comparable SR to GPT-4 and even surpasses it in NE and SPL. 
This demonstrates that our method can adapt effectively to other LLMs. However, the performance of GPT-3.5 is not satisfactory. 
The reason could be that GPT-3.5 extracts less precise constraints than GPT-4 and Claude-3.5.

\vspace{-6pt}
\subsection{Quantitative and Qualitative Analysis}
\textbf{Economic and Low-latency.}
\label{Economic and Responsive}
We compared the response speed and cost efficiency of CA-Nav and NavGPT-CE.
In terms of response speed, we measured the time required for each step to be executed by both methods.
For CA-Nav, this includes the execution time of Grounding-DINO, BLIP2, map building, SLIC, and FMM. 
For NavGPT-CE, this includes the time spent on calling the caption model and the LLM.
By using CSM for minimizing GPT-4 calls, CA-Nav outperforms NavGPT-CE significantly. 
As shown in Figure~\ref{fig:efficient}, NavGPT-CE takes about \textbf{1.29} seconds per action and costs about \textbf{\$0.85} per episode, while CA-Nav responds in \textbf{0.45} seconds and costs only \textbf{\$0.04} per episode, making it roughly 3 times faster and 5$\%$ of the cost.
\vspace{-2em}
\begin{center}
\begin{figure}[h]
\centering
\includegraphics[width=0.45\textwidth,trim=0 10 0 10, clip]{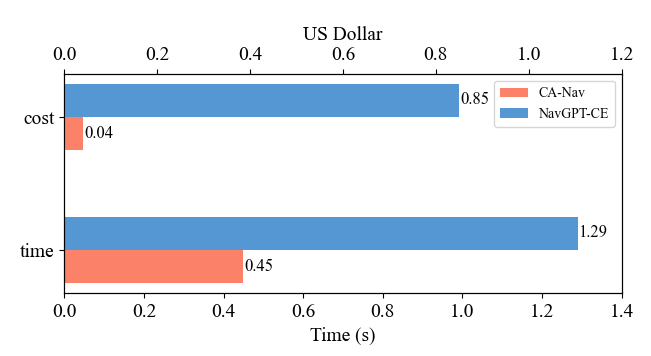}
\caption{Comparison between CA-Nav and NavGPT-CE.}
\label{fig:efficient}
\end{figure}
\vspace{-2em}
\end{center}

\begin{figure}[h]
\centering
\includegraphics[width=0.45\textwidth,trim=10 0 0 10, clip]{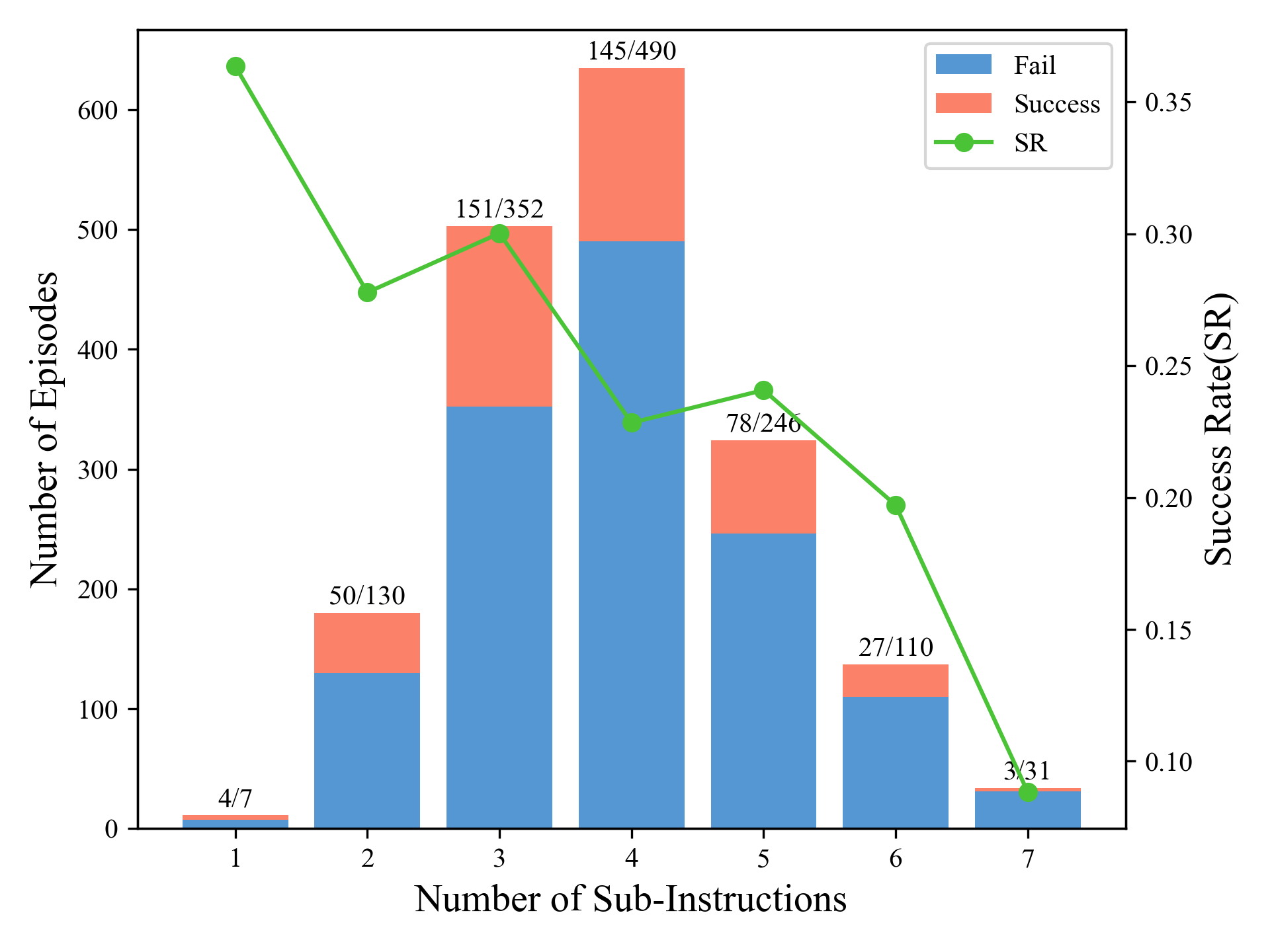}
\caption{Impact of instruction complexity on success rate.}
\label{fig:complex_success_analysis}
\vspace{-1em}
\end{figure}

\vspace{-6pt}
\textbf{Impact of Instruction Complexity.}
To analyze the impact of instruction complexity on navigation performance we divide the instructions of R2R-CE val-unseen split into categories based on the number of sub-instructions, ranging from 1 to 7. 
For each category, we record the number of success and failure trajectories and calculated the corresponding success rate (SR).
The results in Figure~\ref{fig:complex_success_analysis} reveal a trend that as the number of sub-instructions increases, the success rate (SR) generally decreases.
As the number of sub-instructions increases, the system is required to perform more actions and make more decisions, which increases the chance for errors in task execution. 
This becomes particularly apparent in trajectories with 5 or more sub-instructions, where the SR starts to decrease significantly.
This indicates that instruction switching may not always be accurate as the perception model is imperfect.
When the number of sub-instructions increases errors tend to accumulate.
Enhancing the CSM further could be a promising direction for improvement.

\begin{figure*}[ht]
\vspace{-2em}
  \centering
  \includegraphics[width=0.9\textwidth, trim=0 60 0 0, clip]{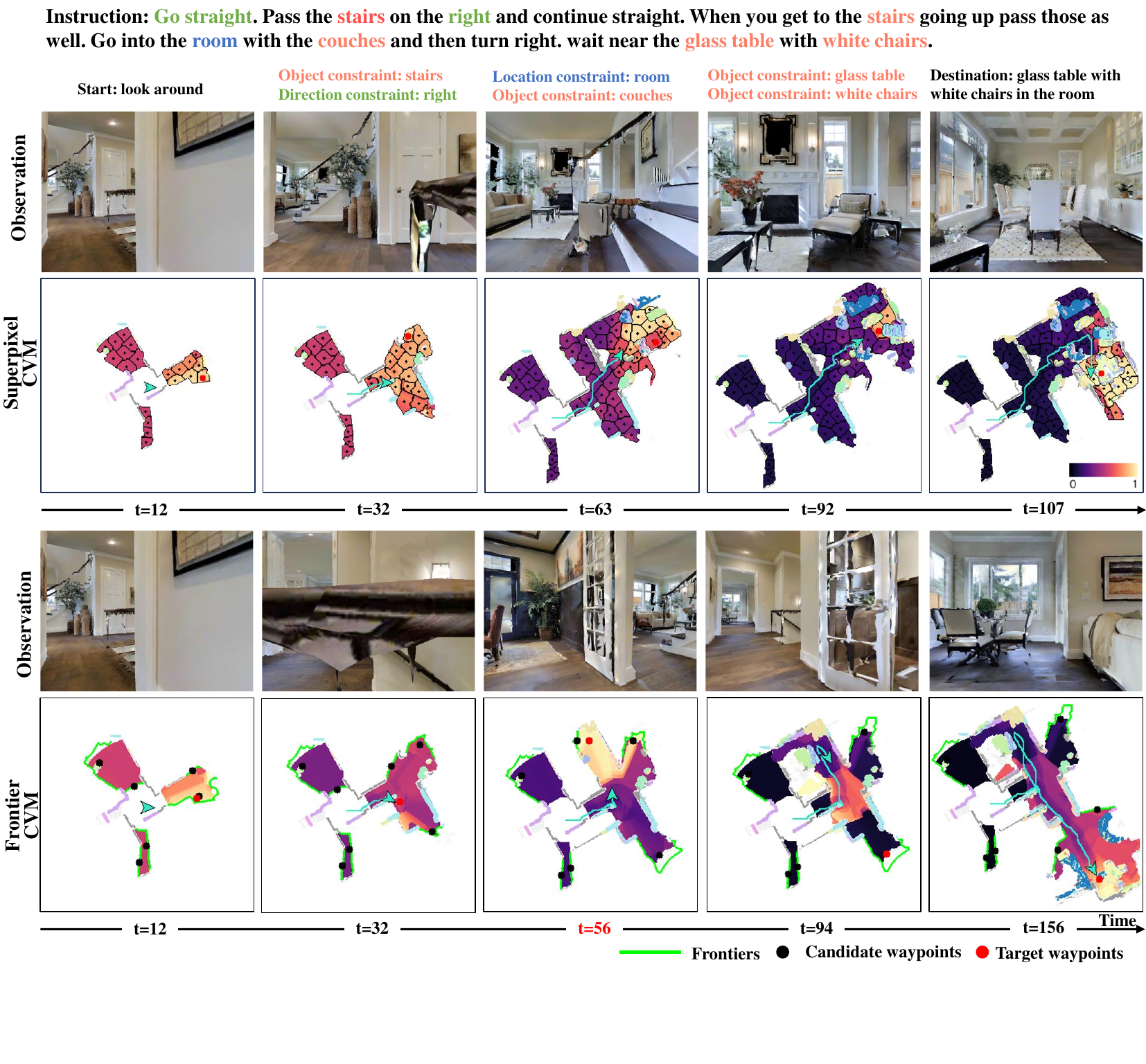}
  \caption{Navigation visualization based on superpixel value map and frontier value map waypoint selection methods.}
  \label{fig:ca-nav_svm_fbe}
\end{figure*}
\vspace{6pt}
\textbf{Constraint Fulfillment Evaluation.}
To demonstrate the effectiveness of the CSM module, we evaluate the fulfillment of the current constraints through human evaluation.
We select the top 25 trajectories from the successful set based on SPL in descending order to assess whether the episodes that best follow the instructions also achieve the highest constraint fulfillment.
Additionally, we randomly sample 25 trajectories from the oracle successful set to examine whether constraint fulfillment declines in episodes that reach the ground truth destination but fail to stop correctly.
For failure trajectories, to ensure a diverse coverage of different trajectory lengths, we randomly select 25 samples each for short, medium, and long trajectories.
Then we calculate the \textbf{C}onstraint \textbf{F}ulfillment \textbf{S}core (i.e., CFS) of episodes and CFS for the three type of constraints (D: Direction, O: Object, L: Location) following equation~\ref{eq:CFS}.
Specifically, $M$ represents the number of the evaluation set which is 25, $N_j$ is the number of constraints in the $j$th episode and $C_j$ represents the number of specific constraints of type direction, object or location in the $j$th episode.
We manually inspected each step of these trajectories and their constraint fulfillment status. 
A step was considered fulfilled if and only if the CSM determined the constraint was met and manual verification confirmed its correctness; otherwise, it was labeled as a violation.
\vspace{-6pt}
{\small
\begin{equation}\label{eq:CFS}
\begin{aligned}
CFS_{\text{episode}}&=\frac{1}{M}\sum^{M}_{j=1}\frac{1}{N_j}\sum^{N_j}_{i=1}\mathbb{I}(\text{constraint}_i)\\
CFS_{D/O/L}&=\frac{1}{M}\sum^{M}_{j=1}\frac{1}{C_j}\sum^{C_j}_{i=1}\mathbb{I}(\text{constraint}_i)
\end{aligned}
\end{equation}}
The results are shown in Table~\ref{failure analysis}.
Successful trajectories achieve the highest episode-level CFS (0.69) and lead in CFS across all constraint types, demonstrating the effectiveness of CSM in adhering to constraints. 
In contrast, Oracle Success trajectories exhibit a lower $\text{CFS}_{episode}$ (0.50) and reduced CFS for individual constraint types, suggesting that while the agent reaches the correct location, it follows constraints less rigorously. 
As failure severity increases, both $\text{CFS}_{episode}$ and $\text{CFS}_{D/O/L}$ drop below 0.5, indicating that CSM fails most of the time.
Overall, these results highlight the relationship between CSM's effectiveness and navigation performance.  
Episodes with higher navigation success rates tend to exhibit greater constraint fulfillment, and this positive correlation to some extent indicates the effectiveness of CSM.

\vspace{-1em}
\begin{center}
\captionof{table}{Constraint Fulfillment Evaluation.}
\vspace{-6pt}
\label{failure analysis}
\small
\resizebox{\columnwidth}{!}{
\begin{tabular}{p{2cm} >{\centering\arraybackslash}p{1.5cm} >{\centering\arraybackslash}p{1.2cm} >{\centering\arraybackslash}p{1.2cm} >{\centering\arraybackslash}p{1.2cm}}
\toprule
Type & $\text{CFS}_{\text{episode}}$ $\uparrow$ & $\text{CFS}_D$ $\uparrow$ & $\text{CFS}_O$ $\uparrow$ & $\text{CFS}_L$ $\uparrow$\\
\midrule
Success & 0.69 & 1.00 & 0.77 & 0.52 \\
Oracle Success & 0.50 & 0.43 & 0.50 & 0.55 \\
Fail(short) & 0.42 & 0.61 & 0.42 & 0.23 \\
Fail(medium) & 0.38 & 0.40 & 0.42 & 0.36 \\
Fail(long) & 0.34 & 0.31 & 0.39 & 0.31 \\
\bottomrule
\end{tabular}}
\end{center}

\begin{figure*}[ht]
  \centering
  \includegraphics[width=\textwidth, trim=0 0 0 0, clip]{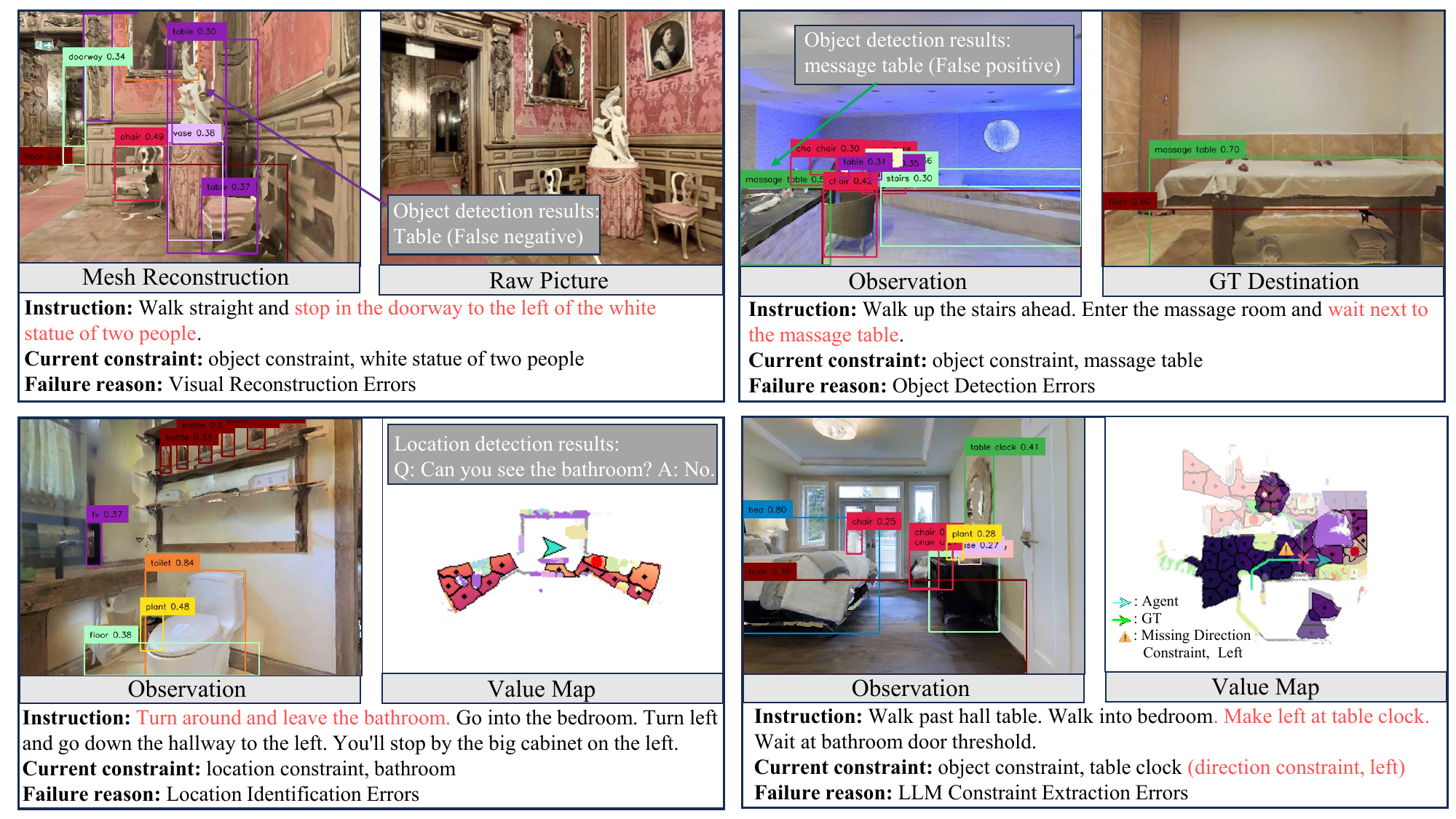}
  \caption{Failure cases visualization. The red font indicates the sub-instruction currently being executed.}
  \label{fig:failure_analysis}
\end{figure*}
\textbf{Navigation Visualization.} 
\label{navigation visualization}
As shown in Figure~\ref{fig:ca-nav_svm_fbe}, and Figure 1 in Appendix, we visualize the navigation process of CA-Nav, FBE-Nav and NavGPT-CE on the same episode. 
Note that the FBE-Nav uses frontier-based exploration for waypoint selection, while the rest of the process is the same as the CA-Nav. 
Among these three methods, CA-Nav shows the most stable and coherent navigation, thanks to CSM's effective instruction decomposition and constraint identification, along with CVM's comprehensive use of the value map. 

In contrast, FBE-Nav fails at step 56 due to abrupt value changes at the frontier. 
As shown in Figure~\ref{fig:ca-nav_svm_fbe}, the agent navigates correctly during the first sub-instruction.
However, upon reaching the second sub-instruction, the landmark prompt changes to ``room," causing the value map, initially based on the previous landmark prompt ``stairs", to decay as new values are updated. 
The FBE-based waypoint selection method then chooses a left frontier, which has the highest value among the available options, but the correct path is to walk straight toward the open area with the stairs.
Compared to the Superpixel Value Map based waypoint selection method shown in Figure~\ref{fig:ca-nav_svm_fbe}, CA-Nav avoids focusing solely on local high values. Instead, it considers the global value map, allowing it to navigate correctly during the second sub-instruction.

\textbf{Failure Analysis.}
To gain deeper insights into the limitations of our approach, we conduct a detailed failure analysis and categorize the four primary failure modes, as illustrated in Figure~\ref{fig:failure_analysis}.
First, visual reconstruction errors in the simulation environment often lead to failures in object detection. Inaccuracies in scene rendering may distort object appearances, causing the detection model to miss or misinterpret key objects.
Second, object detection model errors can arise due to semantic ambiguities. Models such as GroundingDINO may misclassify objects with similar visual and contextual features, leading to incorrect navigation decisions. For example, the bar is mistaken for the message table.
Third, location recognition failures occur when vision-language models such as BLIP2-VQA provide incorrect scene interpretations. Despite the presence of clear visual cues, the model may fail to recognize specific locations. For instance, BLIP2 might not correctly identify a bathroom even when a toilet is visible, resulting in incorrect localization.
Lastly, constraint decomposition errors occasionally arise from the large language model (LLM). In some cases, the LLM fails to accurately parse and decompose instructions, leading to omissions of constraints. For example, the instruction includes a step of ``turn left at table clock", however, the LLM overlooked this sub-instruction and failed to extract the direction constraint. Such errors can disrupt navigation planning and lead to deviations from the intended path.
These failure cases highlight the challenges of integrating multiple vision and language models for embodied navigation and suggest directions for future improvement, such as enhancing simulator's visual fidelity, refining spatial reasoning capabilities, and improving LLM's understanding of instructions.

\subsection{Real Robot Experiments}
\begin{figure}[t]
  \centering
  \includegraphics[width=0.45\textwidth, trim=0 0 0 0, clip]{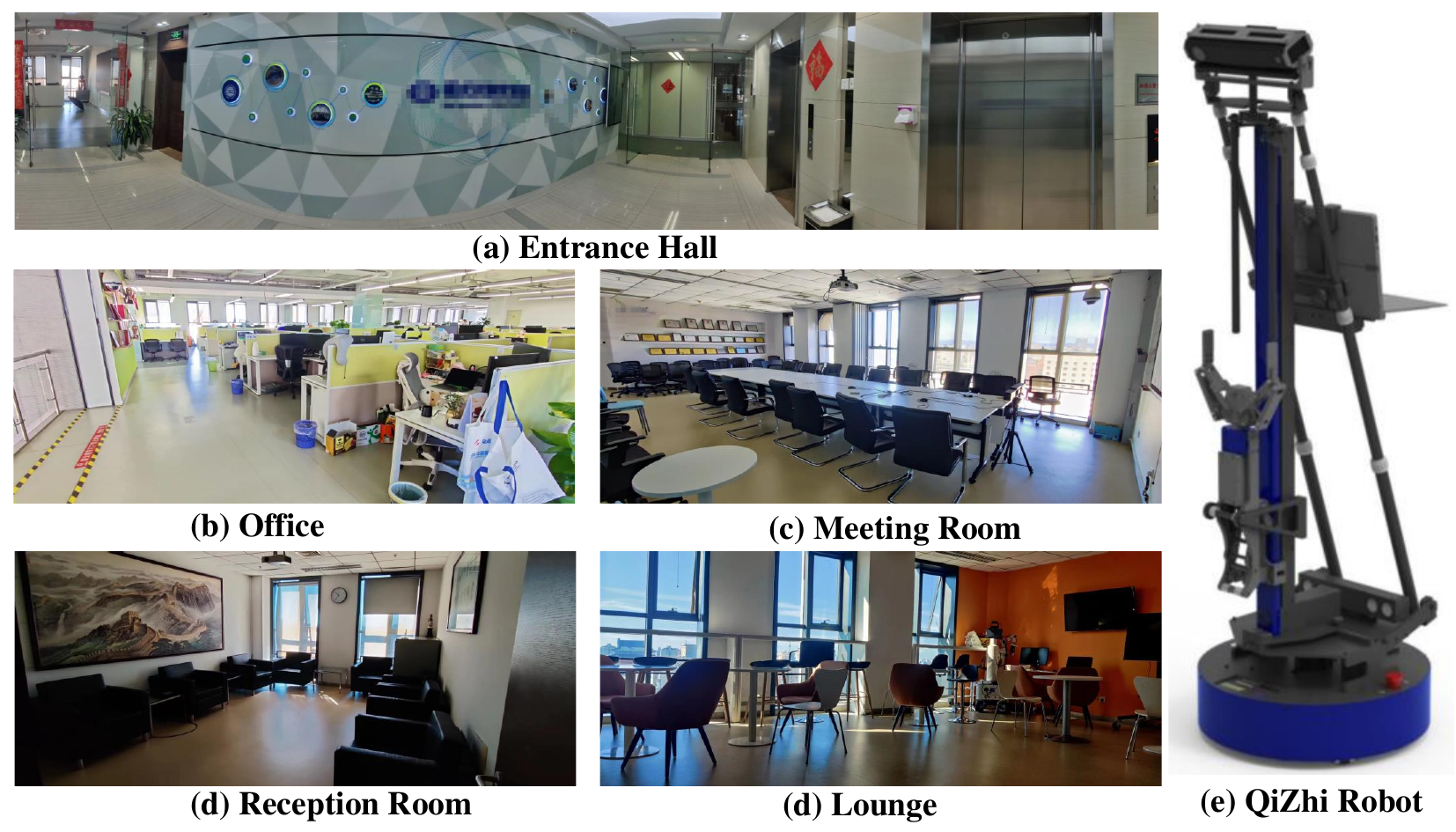}
  \caption{Real-world scenes and real robot.}
  \label{scenes and robot}
  \vspace{-2em}
\end{figure}

We conduct real robot experiments based on the QiZhi mobile robot, as shown in Figure~\ref{scenes and robot} (e).
We equip the robot with a laptop (including an Intel i9-14900HX CPU and a GeForce RTX 4090 GPU) and a Kinect V2.0 RGB-D camera whose HFoV is 84$^\circ$ and VFoV is 42$^\circ$.
However, due to the camera's limited depth sensing range and the inaccuracies or depth loss at the edges, we utilize Depth Anything V2~\cite{depth_anything_v2} to generate depth images.
An RPLIDAR-A2M8 LiDAR is also used to obtain a relatively accurate pose through Hector SLAM. It is important to note that Hector SLAM is employed solely for estimating the camera's pose, not for constructing a pre-built map.
The robot has a radius of 22.5cm and a height of 137cm.

Corresponding to the experiments in simulation (\S~\ref{problem definition}), we still set the low-level action as MOVE FORWARD (0.25m), TURN LEFT/RIGHT (30$^\circ$), and STOP.
The only modification is that successful navigation is defined as the robot stopping within 1 meter of the destination.
For waypoint navigation, we continue to rely on the FMM approach, rather than adopting ROS navigation packages or other trained PointNav policies~\cite{anderson2018spl}.
The linear and angular velocities are set to 0.1m/s and 0.1rad/s respectively.

To evaluate the effectiveness of CA-Nav we conduct experiments in diverse indoor scenes including lounge, meeting room, reception room, entrance hall, and office, as shown in Figure~\ref{scenes and robot}.
These diverse environments test the adaptability of our method to different spatial layouts and semantic contexts.
For instructions, we design 8 instructions with different complexity.
Easy instructions contain only one sub-instruction and the destination is obvious, such as, ``Go to the door.".
Complex instructions are longer and with more than three constraints and the agent can not directly see the destination from its initial position, thus requiring exploration following the instruction.
To check our method's ability to novel landmarks we design open vocabulary destinations such as ``robot" and ``world cup trophy" (details in Table~\ref{tab:real_robot_experiments}).

For each instruction, we run 10 episodes, with the robot's initial pose different each time (details in Table~\ref{tab:real_robot_experiments}).
We deploy CoW-OWL as the baseline. 
For CA-Nav, we use Depth Anything V2 as the default setting and also compare it with ZoeDepth~\cite{bhat2023zoedepth}.
The results indicate that CoW can handle simple cases where the target object is directly visible from the initial pose and the instruction consists of only a single sub-step. 
However, when the instruction requires sequential exploration of multiple objects, CoW fails in most cases.
In contrast, CA-Nav effectively handles long-horizon navigation tasks even without a pre-built map, demonstrating the effectiveness of our real-time constructed constraint-aware value map.
Furthermore, Depth Anything V2 and ZoeDepth exhibit comparable performance. 
However, Depth Anything V2 holds a slight advantage, likely due to its superior depth estimation capability.
In Appendix, we provide visualizations of CoW and ZoeDepth along with corresponding discussions.
Furthermore, thanks to the VLM, CA-Nav demonstrates a certain level of generalization to open vocabularies.
As shown in Figure 2 in Appendix, the agent follows a complex instruction consisting of four sub-instructions and ultimately stops successfully near a World Cup Trophy.
This indicates that CA-Nav can generalize to new instructions and adapt to novel environments while effectively tracking the navigation process.

\setlength{\fboxsep}{1pt}
\begin{table*}[t]
\caption{Information about real robot experiments. The robot's initial pose will be different in each episode. $d$ represents the distance between the initial position and the destination. SR is the success rate. CoW represents the detector based version (i.e., CoW-OWL), Ours represents CA-Nav using Depth Anything V2, ZoeDepth represents CA-Nav using ZoeDepth.}
\vspace{4pt}
\label{tab:real_robot_experiments}
\resizebox{\linewidth}{!}{
	{\renewcommand{\arraystretch}{1.2}
\begin{tabularx}{\textwidth}{>{\centering\arraybackslash}m{3cm} | >{\centering\arraybackslash}m{3.2cm} >{\centering\arraybackslash}m{3.2cm} >{\centering\arraybackslash}m{3.2cm} >{\centering\arraybackslash}m{3.2cm}}
\hline
\textbf{Instruction} & Go to the door. & Turn left then walk towards the signboard and wait by the elevator. & Walk out of the door then stop in front of the plant. & Walk out to the corridor, and stop in front of the poster. \\
\hline
\textbf{Example Pose} & 
\raisebox{-.05\height}{\colorbox{white}{\includegraphics[width=3.5cm]{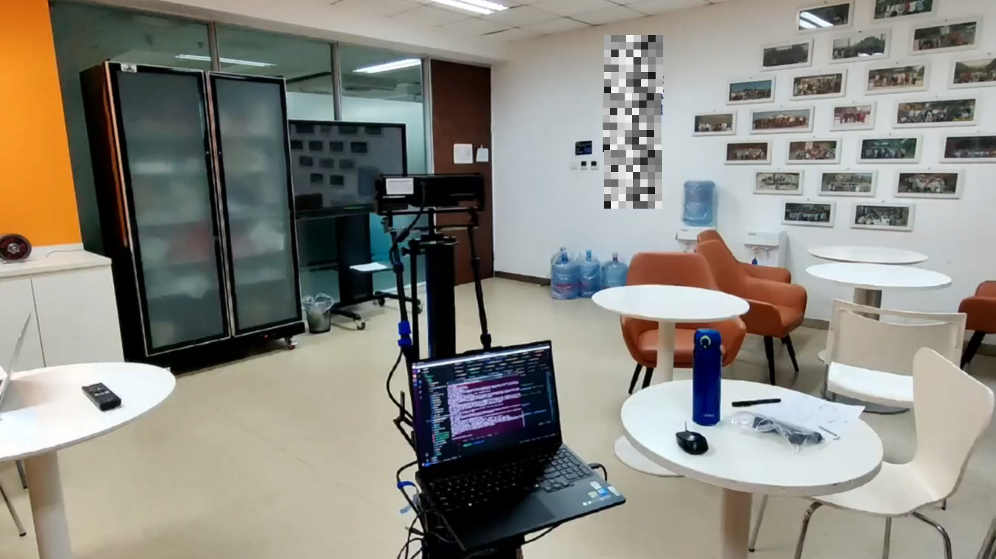}}} & 
\raisebox{-.05\height}{\colorbox{white}{\includegraphics[width=3.5cm]{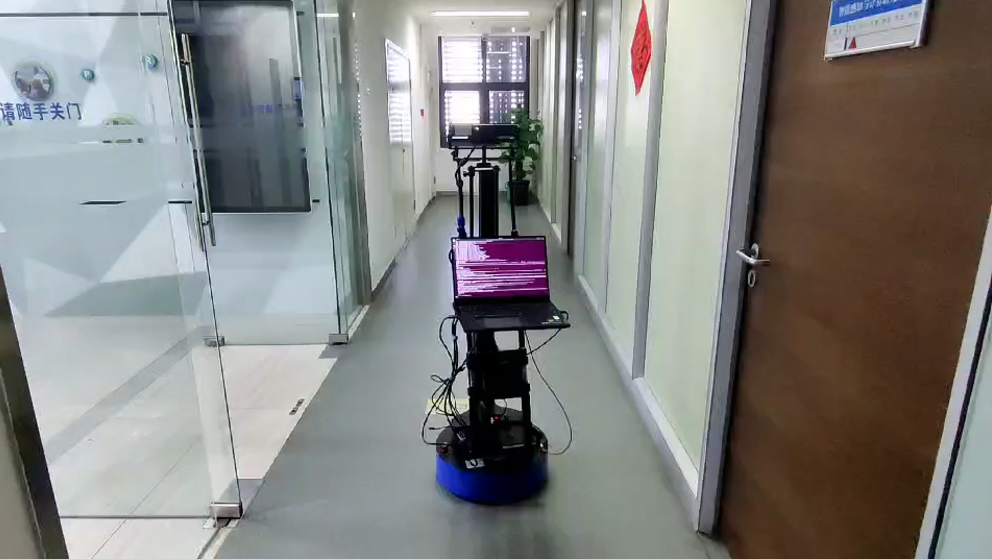}}} & 
\raisebox{-.05\height}{\colorbox{white}{\includegraphics[width=3.5cm]{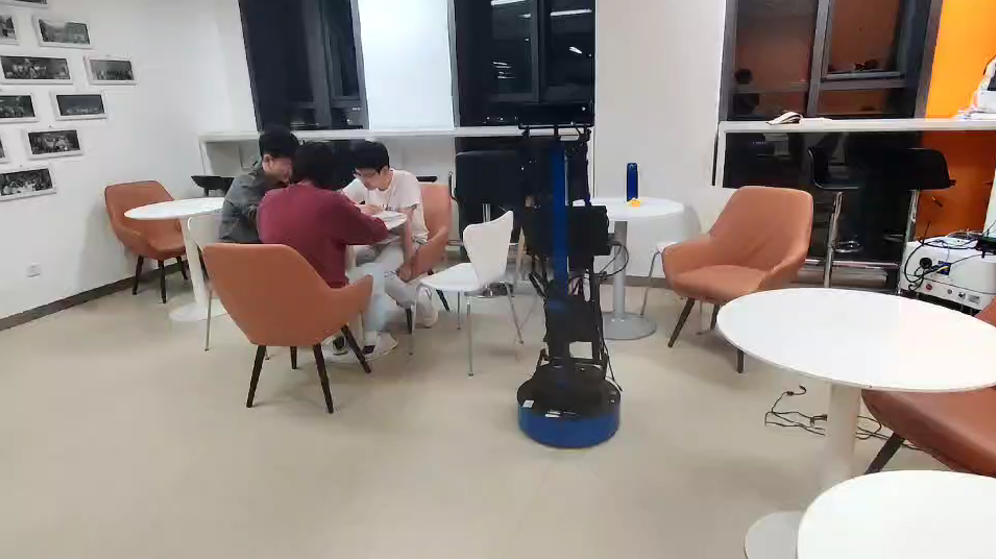}}} & 
\raisebox{-.05\height}{\colorbox{white}{\includegraphics[width=3.5cm]{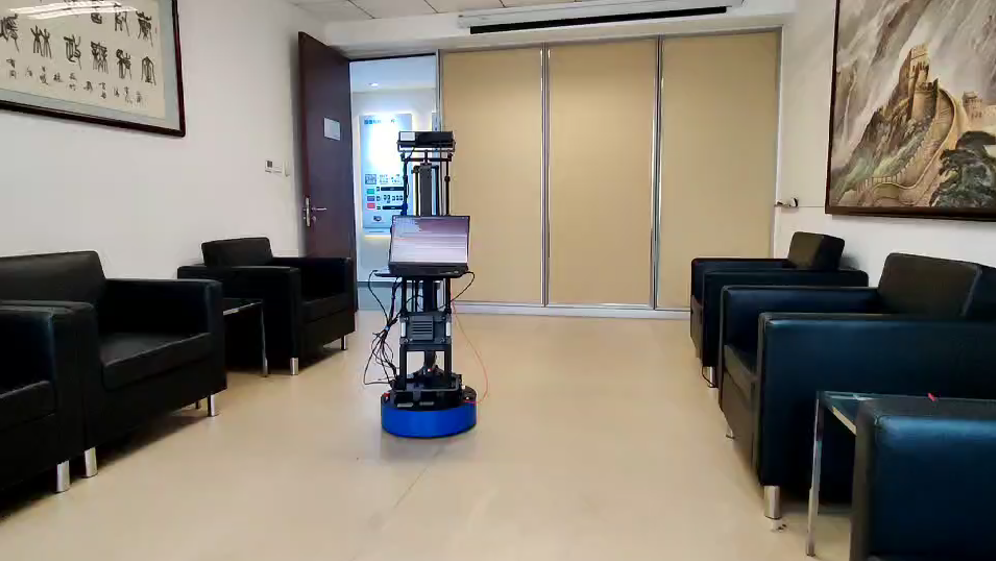}}} \\
\hline
\textbf{Distance ($d$)} & 5.4m & 5.5m & 6.6m & 4.3m \\
\hline
\textbf{SR: CoW, Ours, ZoeD} & 4/10, 8/10, 8/10 & 0/10, 4/10, 3/10 & 1/10, 5/10, 2/10 & 2/10, 6/10, 4/10 \\
\hline
\textbf{Instruction} & Walk towards the robot. & Walk towards the living room then stop beside the couch. & Turn slightly right and walk into the meeting room, step forward then stop in front of the table. & Go towards the plant, turn right, walk along the wall then stop near the World Cup Trophy. \\
\hline
\textbf{Example Pose} & \raisebox{-.05\height}{\colorbox{white}{\includegraphics[width=3.5cm]{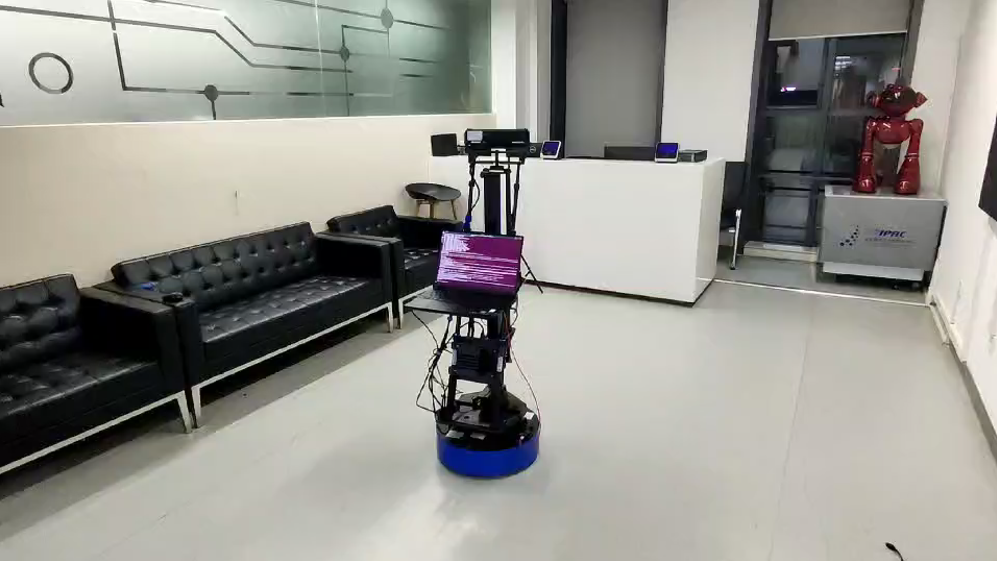}}} & \raisebox{-.05\height}{\colorbox{white}{\includegraphics[width=3.5cm]{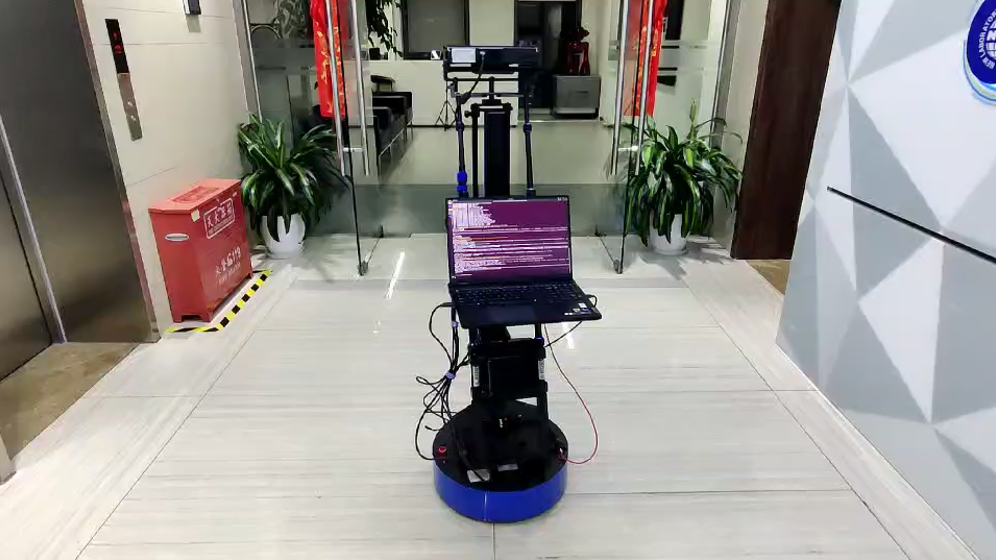}}} & \raisebox{-.05\height}{\colorbox{white}{\includegraphics[width=3.5cm]{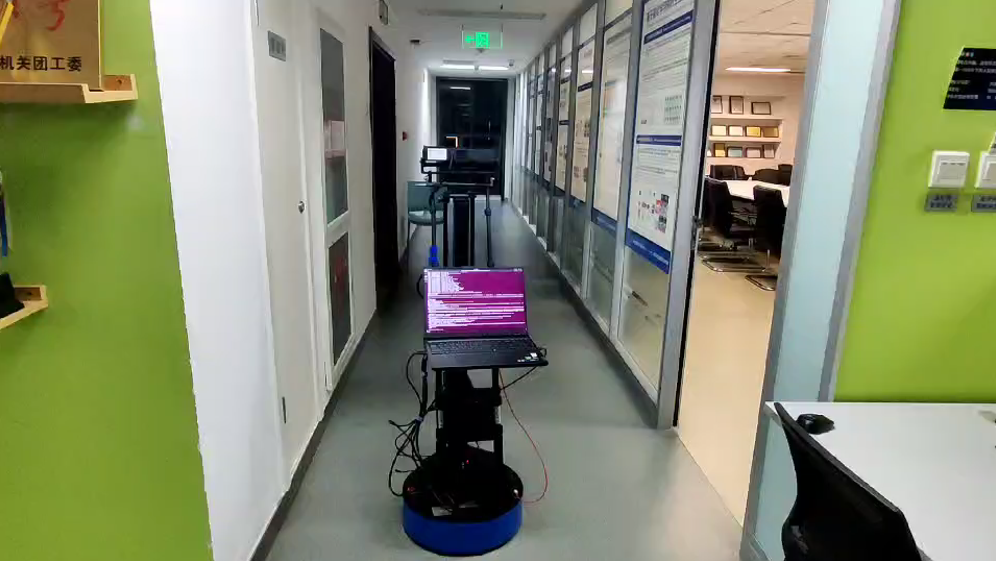}}} & \raisebox{-.05\height}{\colorbox{white}{\includegraphics[width=3.5cm]{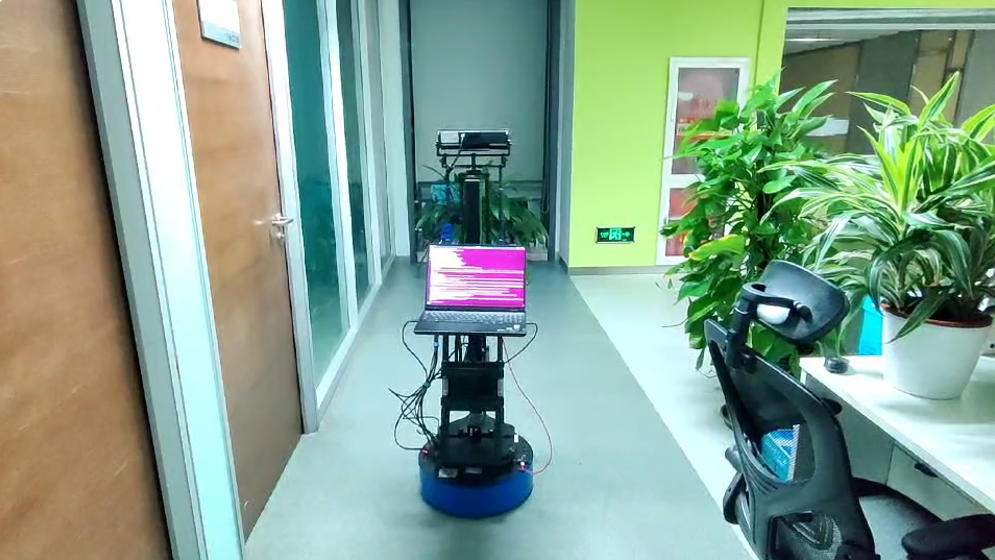}}} \\
\hline
\textbf{Distance ($d$)} & 3.6m & 6.0m & 3.8m & 10.4m \\
\hline
\textbf{SR: CoW, Ours, ZoeD} & 6/10, 7/10, 6/10 & 1/10, 4/10, 3/10 & 0/10, 4/10, 4/10 & 0/10, 2/10, 1/10 \\
\hline
\end{tabularx}}}
\end{table*}
\section{Conclusion}
\label{conclusion}
In this work, we focus on developing a novel Constraint-Aware Navigator for the challenging zero-shot Vision-Language Navigation in Continuous Environments.
To reach this goal, we propose a Constraint-aware Sub-instruction Manager and a Constraint-aware Value Mapper.
The two modules work coherently to navigate novel environments by identifying and adapting to the constraints of each sub-instruction.
Extensive experiments are conducted in both simulated and real-world environments, demonstrating the robustness and adaptability of our approach.
These results highlight the potential of CA-Nav to bridge the gap between simulated benchmarks and real-world applications, providing an effective framework for future advancements in zero-shot navigation tasks.

\textbf{Limitations and Future work.} 
Under the egocentric setting, the agent tends to spend more steps exploring the environment, resulting in slightly lower SPL.
Developing a navigation world model to predict environments outside the current field of view is a potential direction to alleviate this issue.
Moreover, CA-Nav is bottlenecked by the imperfections in perception models, which results in less accurate instruction decomposition and constraint detection, particularly for longer instructions.
Leveraging more robust foundation models and fine-tuning specific modules are promising future directions.

\bibliographystyle{IEEEtran}
\bibliography{reference}

{\appendix
\begin{figure*}[t]
\centering
  \includegraphics[width=0.75\textwidth, trim=0 0 0 0, clip]{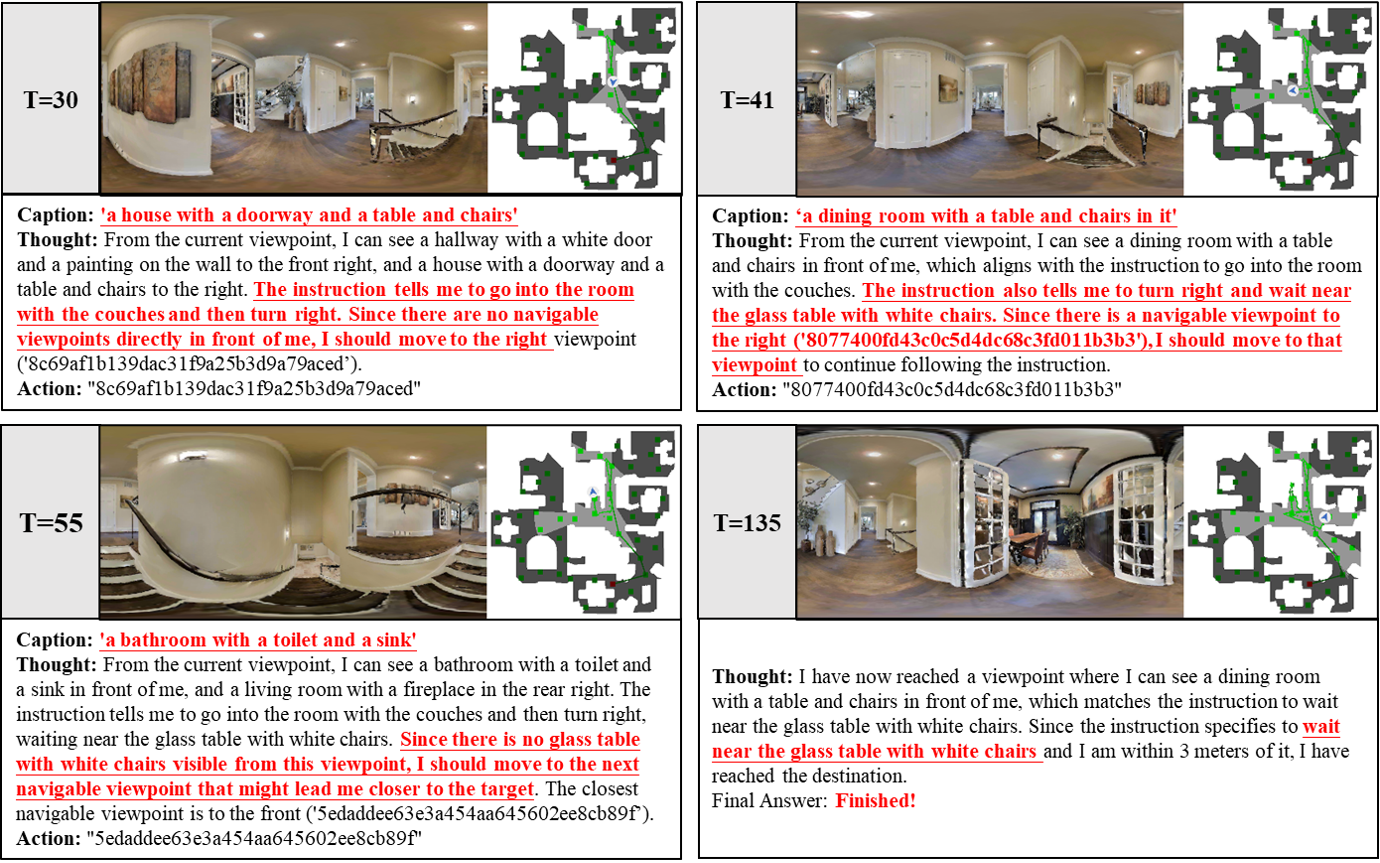}
  \caption{Navigation visualization of NavGPT-CE for the same episode as shown in Figure 8.}
  \label{fig:navgpt-ce}
\end{figure*}

\begin{figure*}[t]
  \centering
  \includegraphics[width=0.8\textwidth, trim=0 70 0 0, clip]{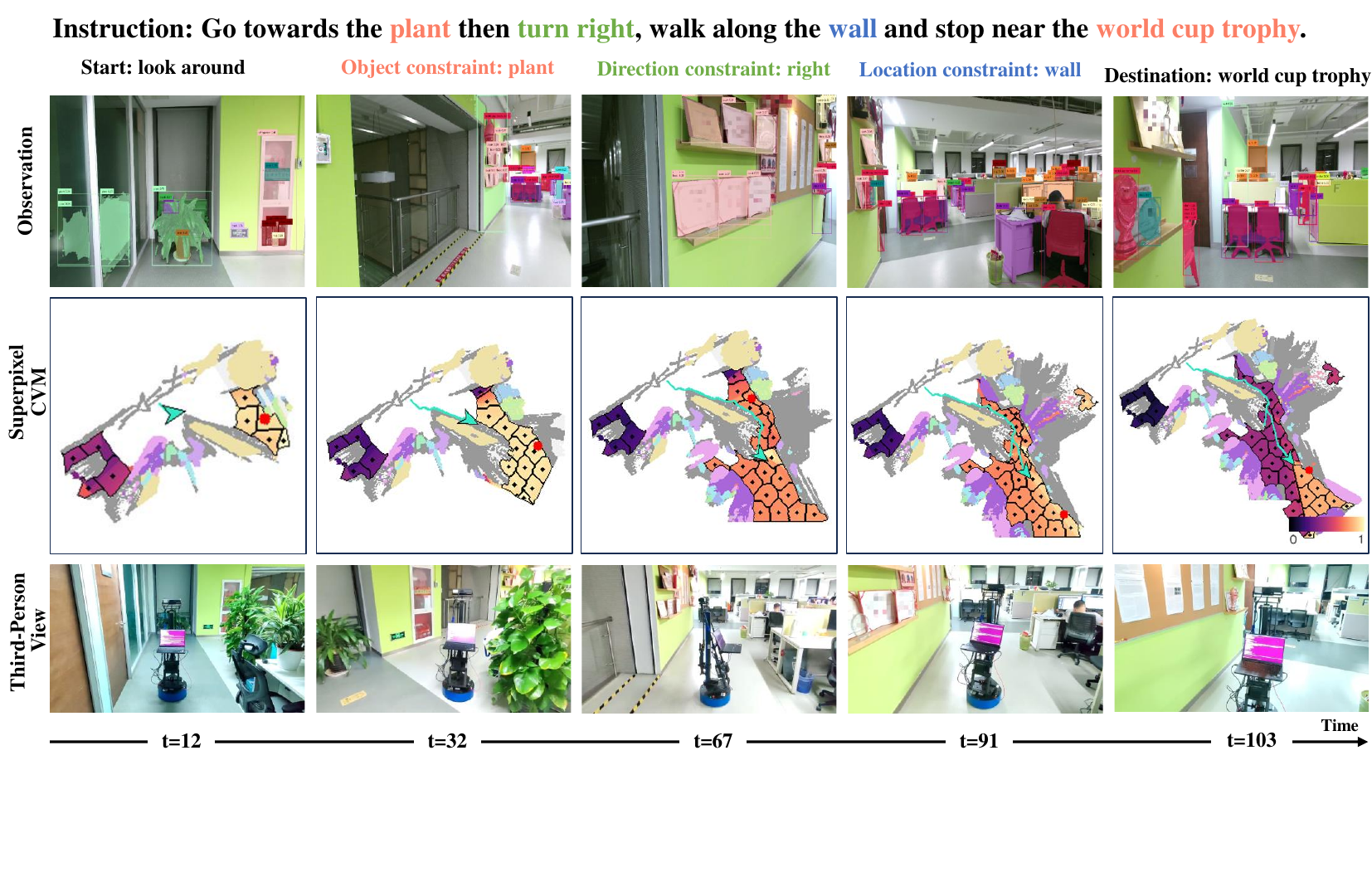}
  \captionsetup{width=0.8\textwidth}
  \caption{Visualization of the real-world navigation. The agent follows a complex instruction consisting
 of four sub-instructions and ultimately stops successfully near a World Cup Trophy.}
  \label{fig:sim2real_trophy}
  \vspace{-16pt}
\end{figure*}
\section*{Visualization and analysis of NavGPT-CE}
We analyze NavGPT-CE in detail and identify three main types of errors that contribute to its failures: waypoint predictor failure, LLM hallucination, and caption model inaccuracy.
As shown in Figure~\ref{fig:navgpt-ce}, the agent initially follows the instructions correctly, but errors start to appear from step 30. 
At step 30, the ground truth action was to move forward, and the LLM correctly identified this.
However, the waypoint predictor failed to identify the navigable viewpoint, causing the agent to move to the right instead. 
At step 41, the LLM exhibits context inconsistency hallucination in summarizing observation descriptions. 
Specifically, the dining room is to the left, which is the correct direction, but the LLM incorrectly summarized it as being in front of the robot. 
Moreover, since the dining room to the left is too far away, the waypoint predictor failed to identify valid waypoints, leading the LLM to choose to go to the right instead.
At step 55, the caption model describes the current panoramic observation as ``a bathroom with a toilet and a sink." 
However, the primary objects in the scene are stairs, and this incorrect caption ultimately misleads the agent.
Our analysis reveals that NavGPT-CE relies solely on converting visual observations into text-based scene descriptions and historical summaries for navigation decisions, making it prone to errors when these descriptions are inaccurate.

\begin{figure*}[t]
  \centering
  \includegraphics[width=\textwidth, trim=0 0 0 0, clip]{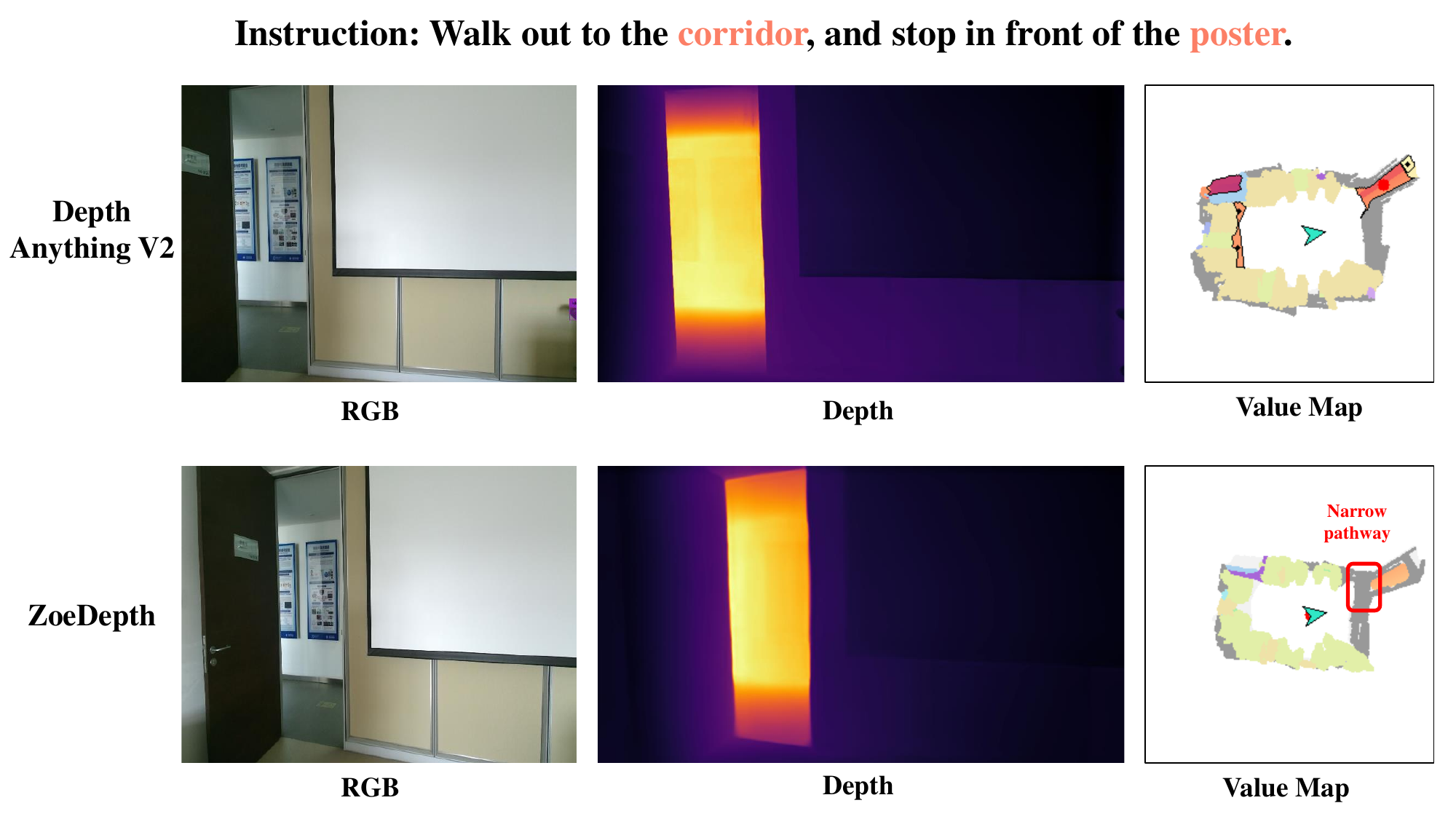}
  \caption{Depth Anything V2 compares with ZoeDepth. As shown in the figure, Depth Anything V2 provides more accurate depth estimation, leading to a more precise value map, particularly in narrow spaces such as pathways. Consequently, Depth Anything V2 achieves a higher success rate than ZoeDepth in real-world experiments.}
  \label{fig:depth-anything-vs-zoedepth}
\end{figure*}

\begin{figure*}[t]
  \centering
  \includegraphics[width=\textwidth, trim=0 0 0 0, clip]{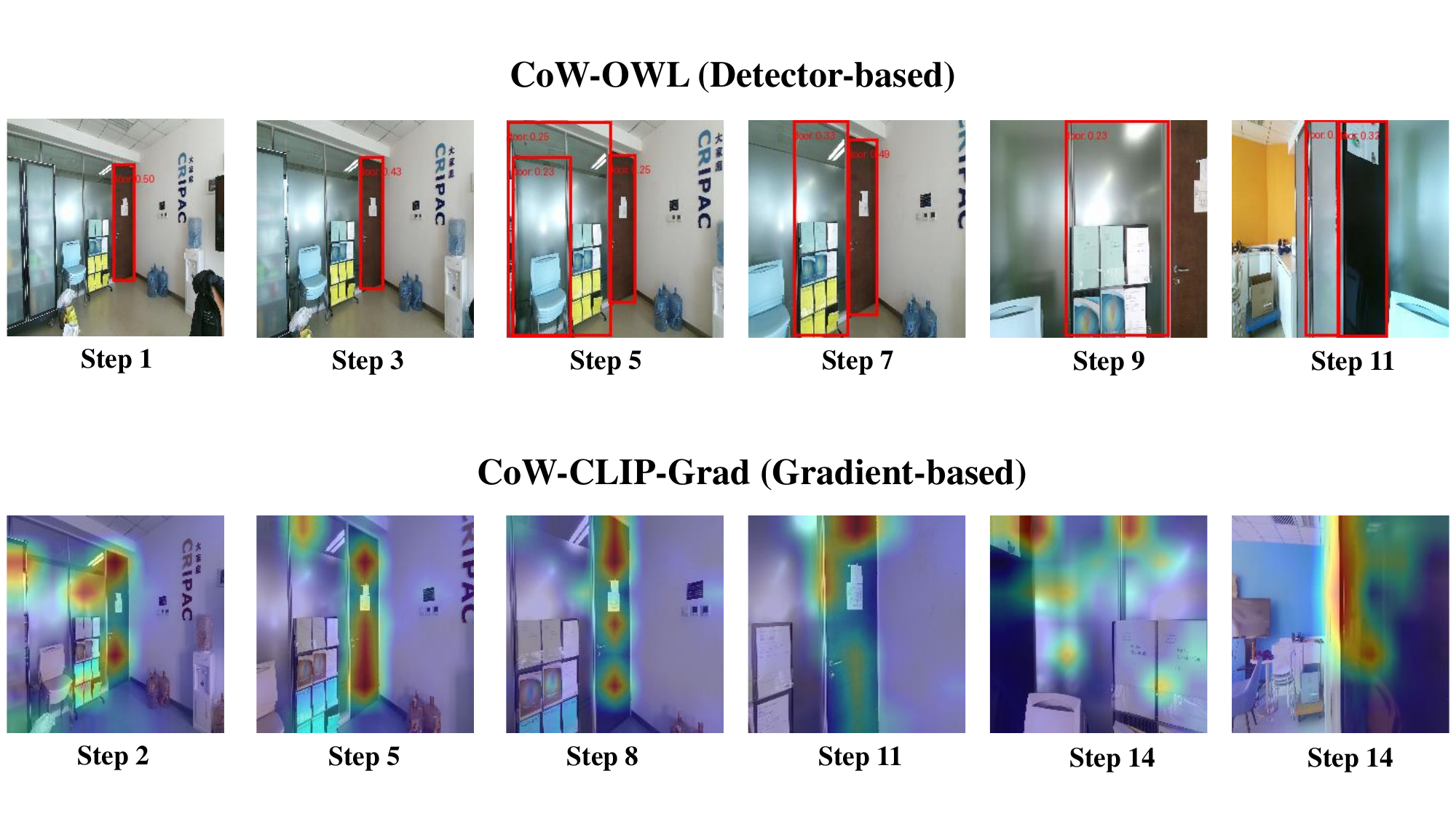}
  \caption{Visualization of the episode with the instruction ``Go to the door." Both CoW-OWL and CoW-CLIP-Grad navigate well initially. However, they become confused when approaching the door due to the presence of a refrigerator door nearby.}
  \label{fig:cow-door}
\end{figure*}

\begin{figure*}[t]
  \centering
  \includegraphics[width=\textwidth, trim=0 0 0 0, clip]{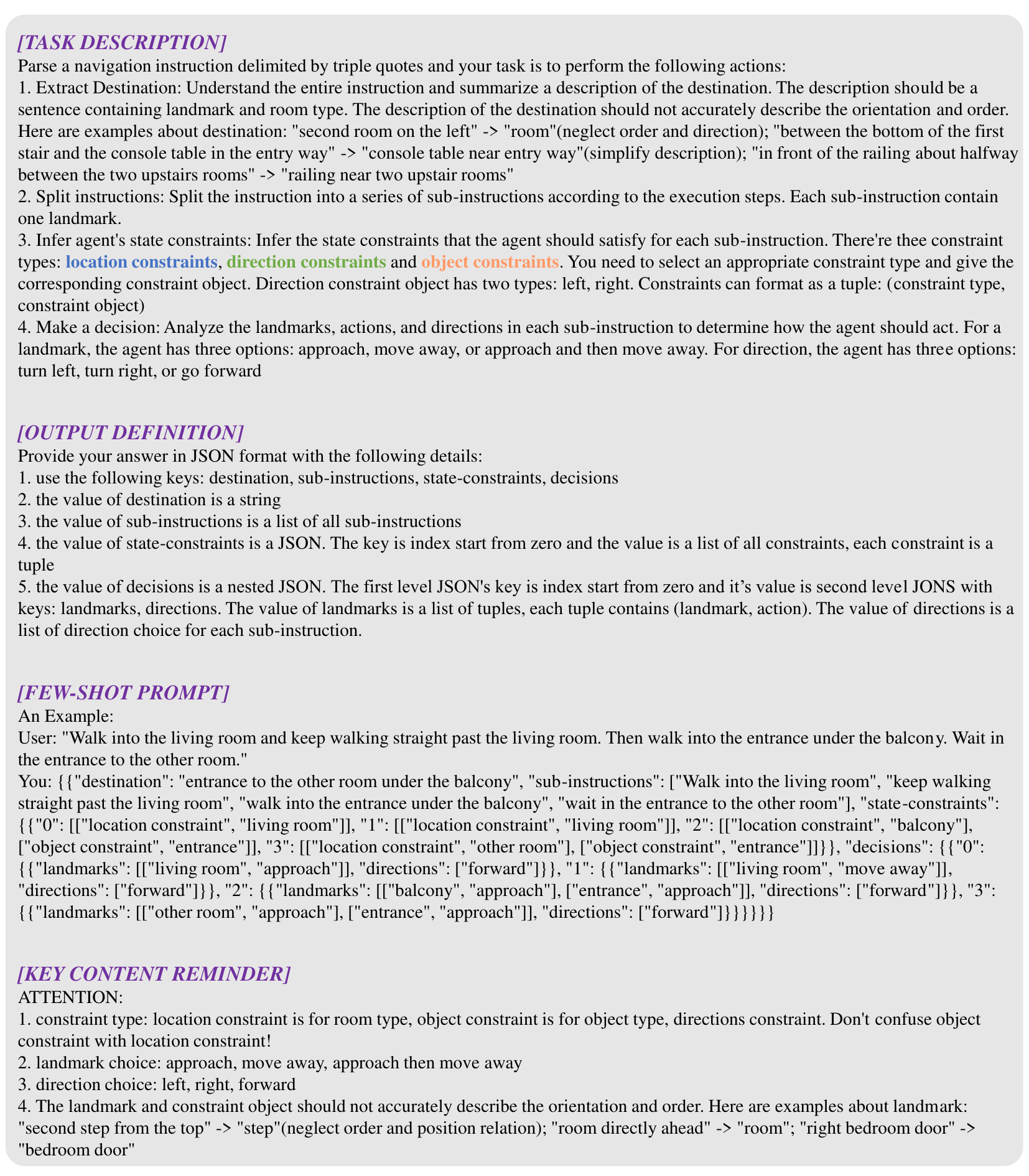}
  \caption{CA-Nav LLM prompts. The prompt consists of four parts, namely task description, output definition, few-shot prompt, and key content reminder.}
  \label{fig:prompts}
\end{figure*}

}

\vfill

\end{document}